\documentclass{article}

\usepackage{PRIMEarxiv}

\usepackage[utf8]{inputenc} % allow utf-8 input
\usepackage[T1]{fontenc}    % use 8-bit T1 fonts
\usepackage{hyperref}       % hyperlinks
\usepackage{url}            % simple URL typesetting
\usepackage{booktabs}       % professional-quality tables
\usepackage{amsfonts}       % blackboard math symbols
\usepackage{nicefrac}       % compact symbols for 1/2, etc.
\usepackage{microtype}      % microtypography
\usepackage{lipsum}
\usepackage{fancyhdr}       % header
\usepackage{graphicx}       % graphics
\usepackage{caption}        % captions
\usepackage{tabularx}       % table
\usepackage{multirow}       % multirow
\usepackage{amssymb}        % amssymb
\usepackage[section]{placeins} %禁止表格或图片到最后一页，在section以内
\title{Advancements in Visual Language Models \par for Remote Sensing: Datasets, Capabilities, and Enhancement Techniques
}

\author{
  Lijie Tao, Haokui Zhang\thanks{Corresponding author}, Haizhao Jing \\
  School of Cybersecurity \\
  Northwestern Polytechnical University \\
  Xi'an, China\\
   \And
  Yu Liu \\
  Zhejiang Lab \\
  Hangzhou, China\\
  \And
  Dawei Yan\\
  School of Cybersecurity \\
  Northwestern Polytechnical University \\
  Xi'an, China\\
  \And
  Guoting Wei\\
   School of Computer Science and Engineering\\
   Nanjing University of Science and Technology\\
   Nanjing, China\\
   \And
   Xizhe Xue \\
   Department of Aerospace and Geodesy \\
   Technical University of Munich\\
   Munich, Germany \\
}

\begin{document}
\maketitle

\begin{abstract}
%\lipsum[1]
Recently, the remarkable success of ChatGPT has sparked a renewed wave of interest in artificial intelligence (AI), and the advancements in visual language models (VLMs) have pushed this enthusiasm to new heights. Differring from previous AI approaches that generally formulated different tasks as discriminative models, VLMs frame tasks as generative models and align language with visual information, enabling the handling of more challenging problems. The remote sensing (RS) field, a highly practical domain, has also embraced this new trend and introduced several VLM-based RS methods that have demonstrated promising performance and enormous potential. In this paper, we first review the fundamental theories related to VLM, then summarize the datasets constructed for VLMs in remote sensing and the various tasks they addressed. Finally, we categorize the improvement methods into three main parts according to the core components of VLMs and provide a detailed introduction and comparison of these methods. A project associated with this review has been created at \href{https://github.com/taolijie11111/VLMs-in-RS-review}{https://github.com/taolijie11111/VLMs-in-RS-review}.
\end{abstract}

% keywords can be removed

\section{Introduction}
Remote sensing technology can help people obtain various observation data from a distance, and it is widely applied in fields such as disaster monitoring~\cite{disaster-monitoring}, urban planning~\cite{urban-planning}, agricultural management~\cite{Tan2023}, surface observation~\cite{surface-observation}, environmental monitoring~\cite{environmental-monitoring},  etc. Consequently, remote sensing image processing technology has been a key topic of focus for researchers both domestically and internationally. Over the past decade, with the development of artificial intelligence technology, significant breakthroughs have also been made in remote sensing image processing techniques. Specifically, various 3D convolutional neural networks (CNNs)~\cite{3DCNN} have been specifically designed for hyperspectral and multispectral image classification, as well as spectral reconstruction. U-Net~\cite{U-Net} is utilized for semantic segmentation and land cover mapping in aerial imagery. A pyramid network has been proposed for multi-scale object detection in Special Administrative Region (SAR) images. The YOLO framework~\cite{YOLO} is applied for small target detection in infrared remote sensing images. Additionally, AI techniques are also applied in fields such as denoising, fusion, enhancement, and compression of remote sensing images, achieving remarkable results. 
 For specific arts in remote sensing, FFCA-YOLO~\cite{zhang2024ffca} designed three plug-and-play modules, which are used to enhance and fuse features in the network, as well as to be spatial context aware, in order to improve local region perception and global correlation as much as possible without increasing complexity. Zhang et~al.~\cite{10781434} improved YOLOv5's perception and detection by integrating spatial-to-depth (SPD) elements, and the CoTC3 module to utilize contextual information. Liu et~al.~\cite{10777291} proposed the MLPA method, which enhances cross-domain hyperspectral imaging (HSI) classification by aligning multi-level features to improve generalization and discriminative ability in the target domain. LBA-MCNet~\cite{10770163} enhanced ORSI-SOD by integrating localization, balancing, and affinity to improve boundary detection and foreground--background context modeling. MSC-GAN~\cite{10769530} leveraged MSC architecture, GFR modules, and channel enhancement to improve multitemporal cloud removal through efficient feature interaction and fusion.
In summary, AI techniques have significantly advanced the processing of remote sensing images, elevating the capabilities in these fields. However, most of these methods primarily focus on image processing, and the models they employ are discriminative models. There are inherent limitations to this type of technology, such as the inability to incorporate some human common sense, and the trained models can only perform a single vision task. 

Recently, due to advancements in data availability, computational power, and the introduction of the Transformer model~\cite{transformer}, natural language processing (NLP) has made significant breakthroughs. By feeding tens of billions of text tokens and scaling the model to tens of billions of parameters, ChatGPT~\cite{chatgpt} achieved amazing performance in natural language understanding, language translation, text generation, question answering, etc. The impressive success of ChatGPT~\cite{chatgpt} has reignited enthusiasm for AI. Larger and more advanced models continue to emerge, showcasing unprecedented advancements in language understanding through their extensive knowledge and sophisticated reasoning abilities, demonstrating human-like thinking. Specifically, Google released the encoder-only model BERT~\cite{BERT}, which pushed the GLUE~\cite{GLUE} benchmark to 80.4\% and achieved an accuracy of 86.7\% on MultiNLI~\cite{MultiNLI} and set a new record of 93.2 for the F1 score on the SQuAD v1.1~\cite{SQuAD} question-answering test, surpassing human performance by 2.0 points. Meta released LLaMA~\cite{LLaMA}, which contains a series of models for 7 billion parameters to 65~billion parameters, and it can beat GPT-3~\cite{gpt-3} with 1/10 parameters. The development of large language models reveals new possibilities for the advancement of artificial intelligence technology, particularly in generative AI. Compared to discriminative AI, generative AI models approach problem-solving in a more flexible and open manner, allowing for the modeling of a broader range of issues. The development of large language models (LLMs) also offers new insights for addressing the challenges faced by discriminative models in the computer vision field.

The introduction of models such as CLIP~\cite{clip} has effectively bridged the modality gap between images and language, creating opportunities for innovative integration. CLIP~\cite{clip}, with its innovative multimodal encoder--decoder structure utilizing BERT~\cite{BERT} as a text encoder and Vision Transformer (ViT)~\cite{vit}/ResNet~\cite{ResNet} as an image encoder, exemplifies the synergy between images and language vocabulary, leading to improved accuracy in classification and detection tasks through a deep understanding of the relationship between image content and textual vocabulary. GLIP~\cite{glip} further enhances this consistency between language sentences and image content. Furthermore, the emergence of VLM has fundamentally addressed numerous constraints associated with traditional discriminative models in remote sensing image processing, thus catalyzing the transition to AI 2.0. Generally speaking, VLM is a type of model designed to integrate and understand both visual and textual information, enabling tasks such as Image Captioning, Visual Question Answering, and crossmodal retrieval by leveraging the relationships between images and language. Compared to previous discriminative models, VLM offers greater flexibility in modeling various AI tasks, allowing it to handle multiple tasks within a single framework. Additionally, its multimodal perceptual ability brings it closer to human perception. In the past two years, both LLMs and VLMs have developed rapidly, with various stronger models being introduced, such as LLaMA~\cite{LLaMA}, LLaVA~\cite{LLaVA}, and GPT-4~\cite{gpt-4}, among others. The performance of related tasks has also been continually improved. 

Similarly to previous discriminative AI technologies, which have significantly improved the field of remote sensing data processing, VLMs—one of the core technologies in AI 2.0—also bring new opportunities for this domain. VLMs, by leveraging their multimodal capabilities, have demonstrated significant progress across various tasks within remote sensing, including geophysical classification~\cite{vlmrscls1,remoteclip}, object detection~\cite{geochat,rs5mgeorscliplargescale}, and scene understanding~\cite{skyscript,wang2023metasegnet}. By introducing LLaVA-1.5~\cite{LLaVA} into the remote sensing field, Kuckreja et~al. proposed GeoChat~\cite{geochat}, which not only answers image-level queries but also accepts region inputs for region-specific dialogue~\cite{geochat}. Deng et~al. designed ChangeChat, which utilizes multimodal instruction tuning to handle complex queries such as change captioning, category-specific quantification, and change localization~\cite{deng2024changechat}. Inspired by GPT-4~\cite{gpt-4}, Hu et~al. developed a high-quality Remote Sensing Image Captioning dataset (RSICap) to facilitate the advancement of large VLMs in the remote sensing domain~\cite{rsgpt}.
Wang et~al.~\cite{add4} proposed a zero-shot target recognition framework for SAR, which combines a vision--language model with a diffusion generative model to achieve the construction of 3D models.
Changen2~\cite{Changen2} is a generative model that creates time-series images with semantic and changeable labels from single-temporal images for collecting and annotating large multitemporal remote sensing datasets.

\begin{figure*}[t]
    \centering
	\includegraphics[width=0.6\textwidth]{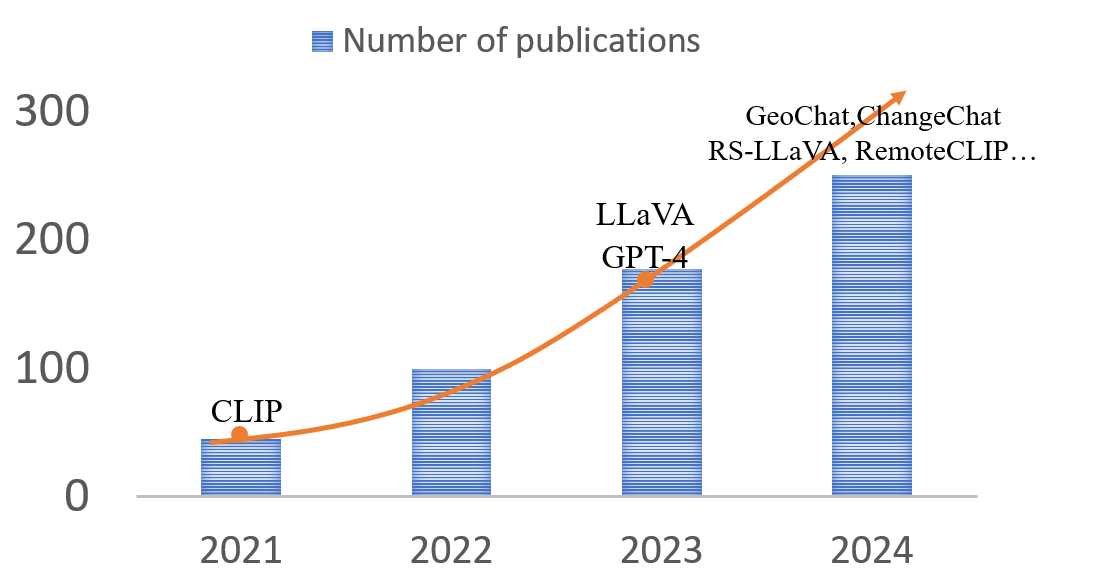}  
    \captionsetup{justification=centering}
	\caption{Number of publications for VLMs in RS (from Web of Science). }
	\label{fig:VLM_evolution}
\end{figure*}

The publication growth trend in Figure~\ref{fig:VLM_evolution} illustrates the development timeline of Visual--Language Models (VLMs) in the remote sensing (RS) field. Growth began after the release of CLIP~\cite{clip}, and in early 2023, following the successive releases of LLaVA \cite{LLaVA} and GPT-4 \cite{gpt-4}, numerous high-performing VLMs started to emerge in the RS domain.

% -----新增application举例 gtwei
In practical applications, VLMs are beginning to demonstrate their potential in the remote sensing domain, albeit still in the early stages of adoption. A prominent example is GeoForge~\cite{GeoForge}, which utilizes a customized GPT-4 model trained with additional geospatial vocabulary and knowledge to handle tasks like mapping administrative boundaries and geocoding major cities. For instance, users can interact with GeoForge to easily obtain the administrative boundaries of Bangladesh and the names of its five most important cities, and can further generate geocoded markers for these five cities based on the obtained information. Despite these advancements, the range of practical applications remains limited, and widespread industry implementation is yet to be realized. This early stage underscores the significant potential of VLMs like GeoForge to enhance remote sensing workflows while highlighting the need for ongoing development to overcome existing challenges and expand their real-world applicability.

% -----新增VLMs对长尾分布/多任务输出的潜力探索(审稿人4 comment 5) gtwei
Beyond these direct applications, VLMs also demonstrate potential advantages over traditional models in handling long-tail distributions and supporting multi-task learning. Pre-trained on extensive and diverse datasets, VLMs can achieve strong generalization capabilities, enabling them to effectively generalize to rare categories and thus mitigate the challenges associated with long-tail distributions. In addition, thanks to their generative output paradigm, VLMs can simultaneously tackle multiple tasks—such as classification, question answering, and counting—within a single framework.
% -----新增VLMs对长尾分布/多任务输出的潜力探索 gtwei

In this paper, we present a comprehensive review of VLM-based remote sensing applications; to ensure coverage of relevant research, we utilized Google Scholar and employed keywords such as “Visual language models for remote sensing”, “Multimodal remote sensing datasets”, and “Deep learning for remote sensing imagery”, offering a thorough delineation of the principal methodologies and the most recent advancements in VLMs, particularly within the context of remote sensing.
% ---增加关键词描述 gtwei
By providing an in-depth introduction to the most representative and influential works in this domain, the survey enables researchers to quickly acquire a nuanced understanding of the development trajectory of these methodologies, facilitating the classification and evaluation of different approaches. Furthermore, the survey extends its contribution by presenting a detailed classification of enhancement techniques, focusing on key components such as vision encoders, text encoders, and alignment between vision and language. This classification, which differs from existing surveys categorized by task type~\cite{survey1}, not only systematically illustrates the research progress made within this specialized field but also serves as a valuable reference point, supporting and inspiring further innovative research endeavors by scholars aiming to explore and advance the boundaries of VLM applications in remote sensing.

\begin{figure*}[t]
    \centering
	\includegraphics[width=0.6\textwidth]{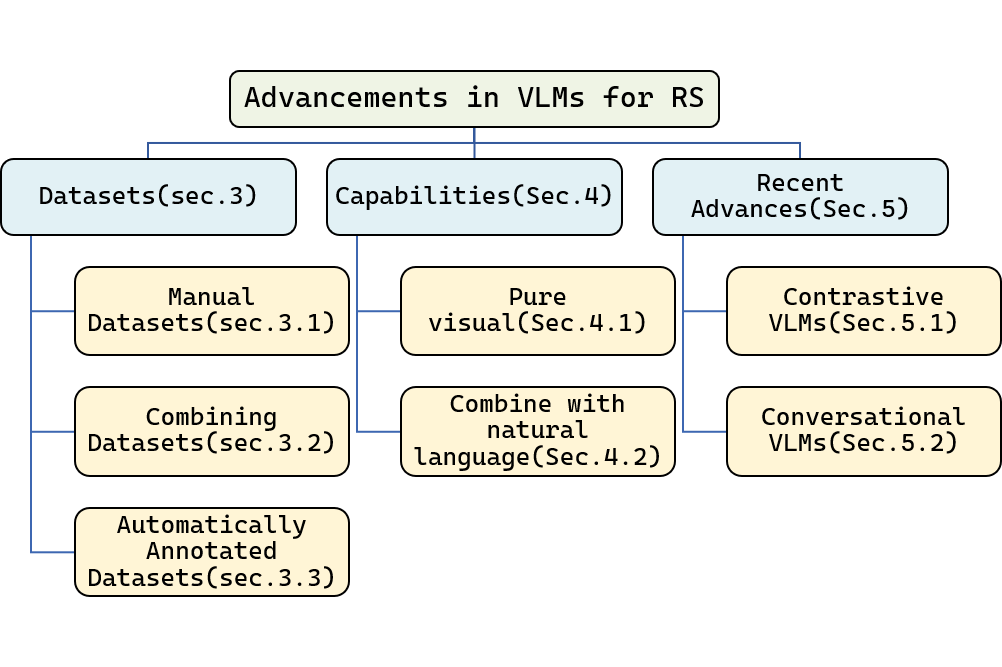}
    \captionsetup{justification=centering}
	\caption{Organization of this survey.}
	\label{fig:organization}
\end{figure*}

%\textbf{\hl{Survey Pipeline} %MDPI: please confirm if it redundant and can be removed, or if it should be heading 1.2} 
This survey is organized as shown in Figure~\ref{fig:organization}. Section~\ref{sec:Foundation Models} presents a convincing introduction to foundation models, including Transformer, Vision Transformer, and the famous VLM—LLaVA.  Section~\ref{sec:Datasets} outlines the datasets used in VLM for RS, which can be further divided manually into collected data, data combined with existing datasets, and data enhanced by VLMs and LLMs.  Section~\ref{sec:Capabilities} describes the capabilities of VLMs, specifically introducing the pure visual tasks and vision--language tasks. In Section~\ref{sec:Enhancement Techniques}, current VLM-based remote sensing data processing methods are introduced according to three major improvement directions, with a comparison and analysis of different approaches. Finally, several promising future directions are discussed in Section~\ref{sec:con}. 

%------------------------------------------------------------%
\begin{figure*}[t]
    \centering
	\includegraphics[width=0.4\textwidth]{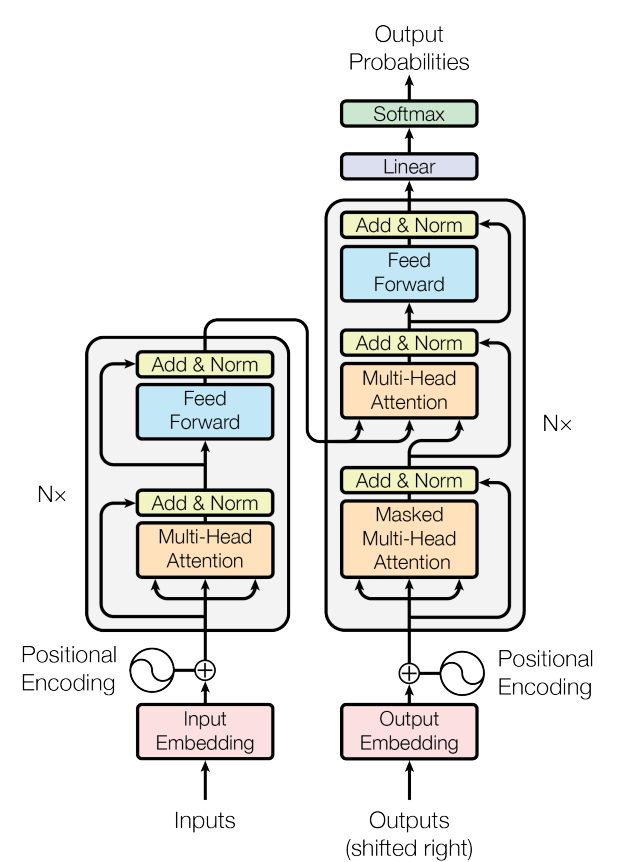}
    \captionsetup{justification=centering}
	\caption{An illustration of Transformer~\cite{transformer}.}
	\label{fig:Transformer}
\end{figure*}

\section{Foundation Models}
\label{sec:Foundation Models}
\subsection{Transformer}
The emergence of the Transformer has changed the paradigm in the NLP field and laid the foundation for the subsequent development of LLMs. As shown in Figure~\ref{fig:Transformer}, the Transformer consists of an encoder and a decoder, each comprising an attention layer (also known as the token mixer layer) and a Feed Forward layer (FFN, sometimes referred to as the channel mixer layer). The attention layer allows for the exchange of information among all input tokens, while the Feed Forward layer facilitates information exchange across channels.The FFN layer consists of two linear layers. The width of the first linear layer is set to 4$\times$ (the scale factor) that of the input tokens, while the second linear layer matches the width of the input tokens. The attention mechanism is formulated as follows:

\begin{equation}
Attention(Q,K,V)=Softmax\left(\frac{QK^T}{\sqrt{d_k}} \right)V
\end{equation}

\noindent where $d_k$ is a scaling factor, typically set to the dimension of the input feature head. $Q$%MDPI: Please keep the format (italic/non-italic) consistent throughout the manuscript. the same as following
, $K$, and $V$ are variations of the input data, usually generated by a linear layer. For instance, $Q=W_q(x)$, $K=W_k(x)$, $V=W_v(x)$, where $x$ denotes the input feature and $W_q$, $W_k$ and $W_v$ are learnable parameters of the linear layers. When Q, K and V are variations of the same input feature,  it is referred to as self-attention. If Q varies from $x_1$ while K and V vary from $x_2$ (with $x_1$ and $x_2$ coming from different sources), it is called cross-attention.

The Transformer model offers several advantages over traditional RNNs and LSTMs. Firstly, it utilizes self-attention mechanisms, allowing it to process input sequences in parallel rather than sequentially, which significantly speeds up training. Secondly, the ability to capture long-range dependencies without the vanishing gradient problem enhances its effectiveness in understanding context. This efficiency and scalability enable the development of large language models (LLMs), as Transformers can handle vast amounts of data and complex patterns more effectively than RNNs and LSTMs, paving the way for advancements in natural language processing tasks.

\subsection{Vision Transformer}

\begin{figure*}[t]
	\centering
	\includegraphics[width=0.8\textwidth]{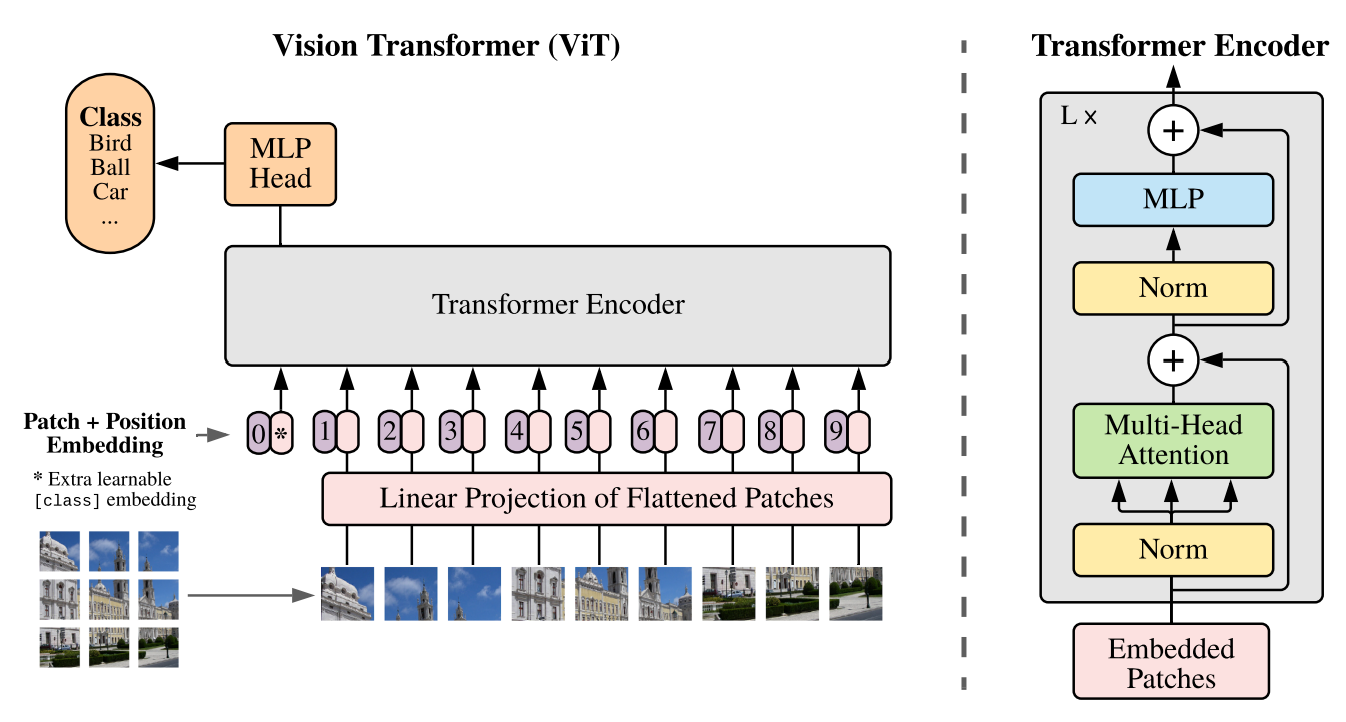}
    \captionsetup{justification=centering}
	\caption{ The illustration of Vision Transformer \cite{vit} }
	\label{fig:ViT}
\end{figure*}

Inspired by the success and popularity of transformer used in NLP, the Vision Transformer (ViT) model was proposed later and  tailored for the computer vision field. As shown in Figure~\ref{fig:ViT}, ViT is composed of three components: Linear Projection of 
Flattened Patches (embedding layer), Transformer Encoder, and MLP Head. In summary, to enable NLP models to process images, the images are serialized by cropping them into uniform-sized patches. These patches are then transformed into features using a projection layer. To ensure the model is sensitive to the positional information of the image patches, position embeddings are added to the corresponding patch embeddings. Currently, there are four types of position, embedding methods, including (1) absolute position embedding; (2)~relative position embedding (RPE)~\cite{RPE}; (3) conditional position embedding (CPE)~\cite{CPE}; (4)~rotary position embedding (RoPE)~\cite{roPE}. The ViT adopts the first one. 

Compared to convolutional networks, ViT has a significant advantage in scaling-up ability. Its performance on image processing does not saturate quickly as the model size increases, allowing for better utilization of larger models and enhanced capabilities in handling more complex tasks. This characteristic has sparked a surge of research interest in the ViT architecture, leading to the continuous emergence of various improved versions of ViT models, such as SwinTransformer~\cite{swintran}, Deit~\cite{deit}, etc. So far, the model paradigms in both the NLP and CV fields have been unified, which has laid the foundation for the emergence of VLMs.  

\subsection{Vision Language Model}

\begin{figure*}[t]
	\centering
	\includegraphics[width=0.6\textwidth]{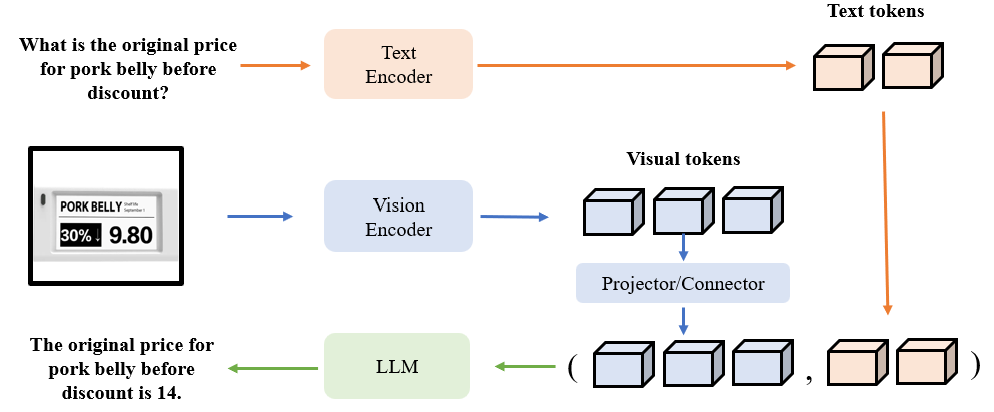}
    \captionsetup{justification=centering}
	\caption{ The illustration of LLaVA \cite{LLaVA} }
	\label{fig:LLaVA}
\end{figure*}

\begin{figure*}[t]
	\centering
	\includegraphics[width=0.6\textwidth]{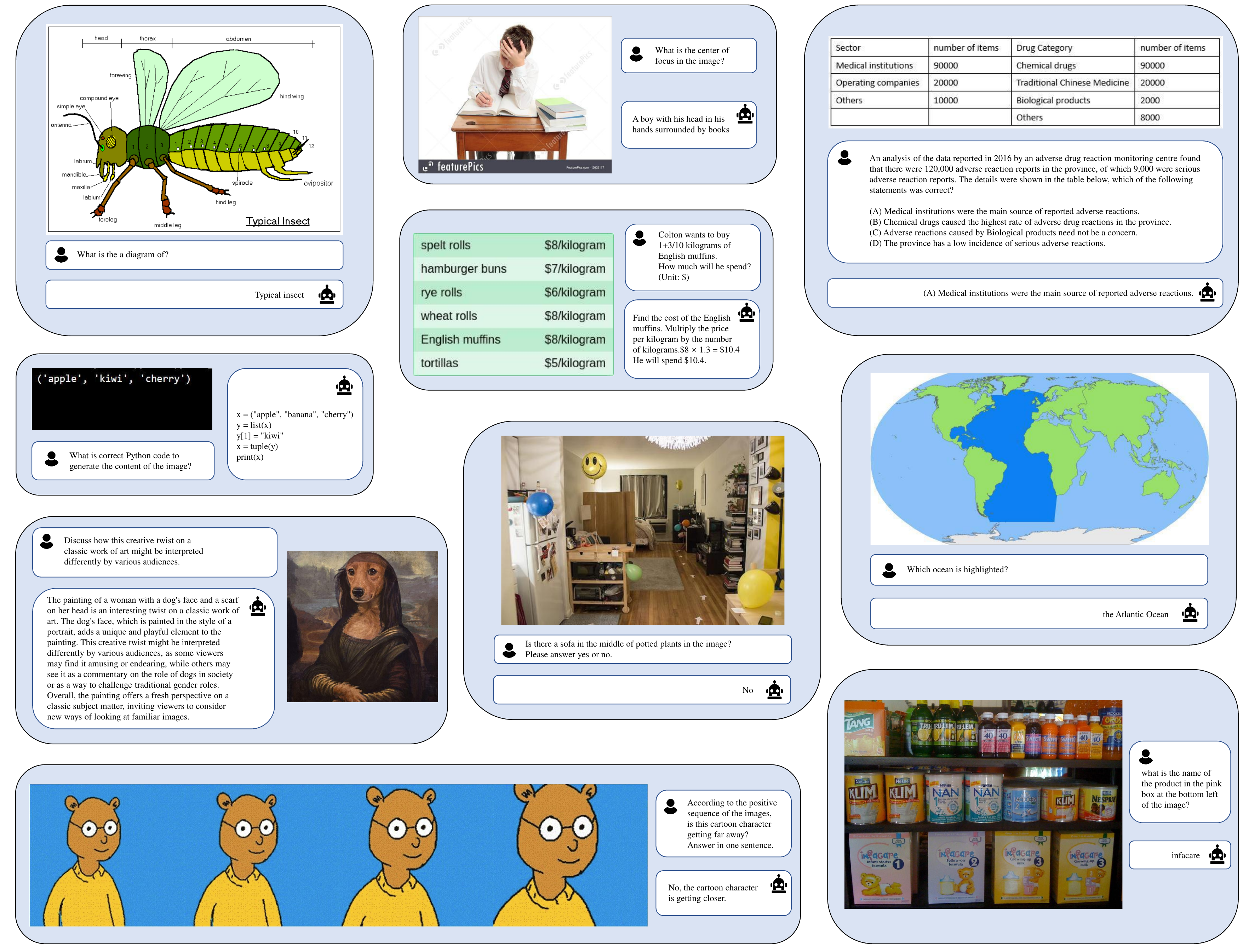}
    \captionsetup{justification=centering}
	\caption{ Ten representative tasks of VLM \cite{yan2024tg} }
	\label{fig:examples}
\end{figure*}
The introduction of Transformers has significantly advanced language processing, while the emergence of ViT has unified processing paradigms in both NLP and CV. Subsequently, CLIP~\cite{clip} breaks down the remaining barriers between language and images. Building on these foundations, VLMs have been proposed by combining ViT and LLMs. A representative paradigm among these VLMs is LLaVA~\cite{LLaVA}, whose architecture is illustrated in Figure~\ref{fig:LLaVA}.

The core idea of LLaVA is to integrate visual tokens with textual tokens and feed them into an LLM to produce text-based answers. Specifically, the visual encoder extracts image features, and a projection layer converts these features into language-embedding tokens, bridging the gap between different modalities. These visual tokens are subsequently input into the LLM together with the text tokens, enabling multimodal interaction.

In practice, a pre-trained CLIP model is often employed as the visual encoder to extract visual tokens, because it has been trained extensively on large-scale image--text pairs. This training allows the CLIP visual encoder to map image features into a representation space comparable to that of text features, laying the groundwork for subsequent conversion into language-embedding tokens.

Although the merging approach is straightforward, the resulting VLM can handle a wider range of tasks and interact with users, enabling conversations based on the content of images. Figure~\ref{fig:examples} illustrates ten representative tasks of VLM. From the results, it is evident that the model can not only perceive the content of images but also possesses a certain degree of reasoning ability, as demonstrated in the example shown in the bottom left corner of the Figure~\ref{fig:examples}.

\section{Datasets}
\label{sec:Datasets}
\begin{figure*}[t]
	\centering
	\includegraphics[width=0.8\textwidth]{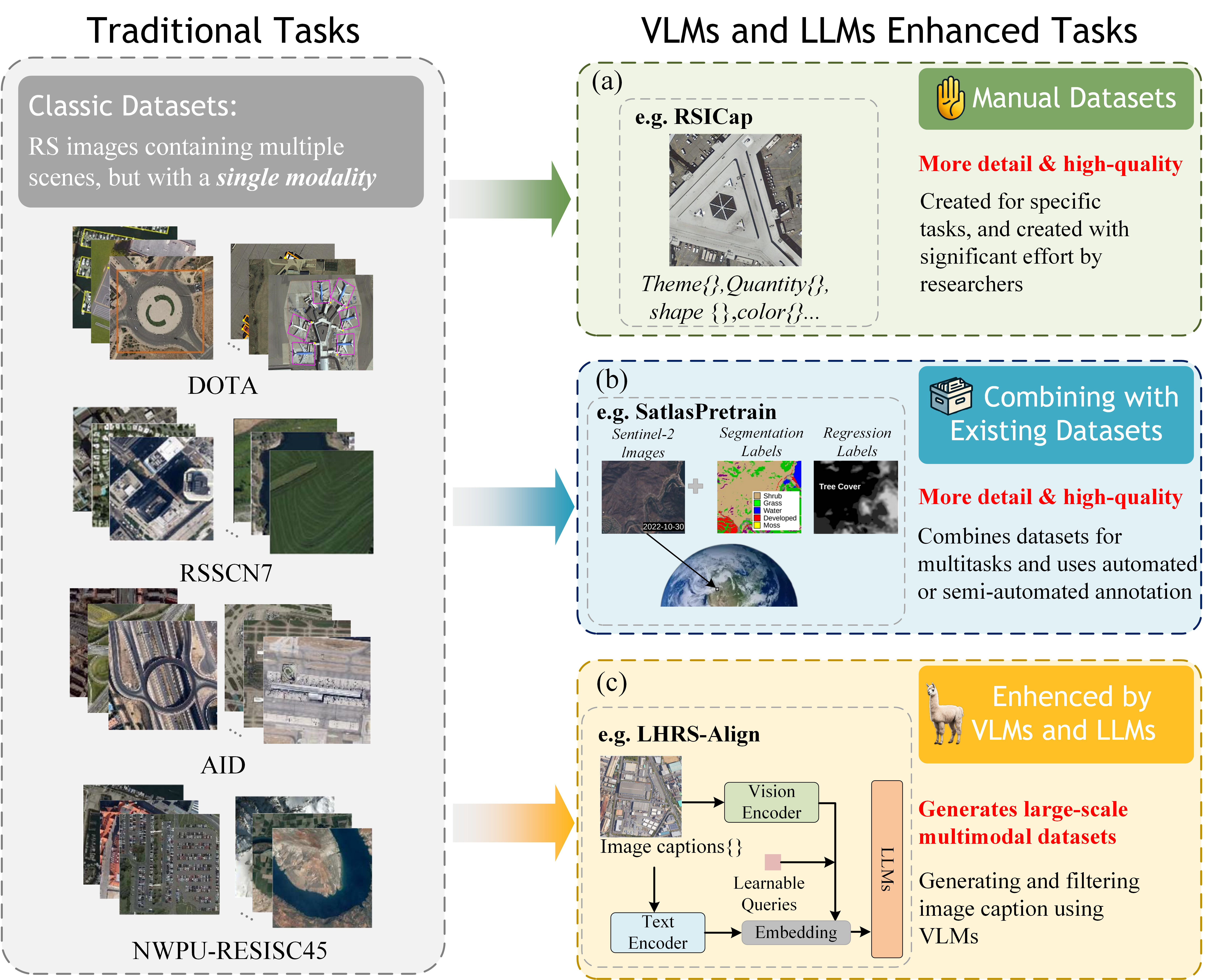}
    \captionsetup{justification=centering}
	\caption{Comparison of traditional (left) and VLM datasets (right) in remote sensing.}
	\label{fig:Datasets}
\end{figure*}

Whether it is the previous discriminative models or the current generative models, they all fall under data-driven models. Therefore, VLM-based remote sensing image processing methods also require corresponding training data as support. Figure~\ref{fig:Datasets} compares the datasets used in previous discriminative models with those used in current VLMs. From the figure, we can observe that the most significant improvement is that VLM datasets include both image and language annotations. 

In the remote sensing field, existing VLM datasets can be broadly categorized into three types based on their production approaches:
\begin{itemize}
    \item Manual datasets, which are completely manually annotated by humans based on specific task requirements.
    \item Combined datasets, which are built by combining several existing datasets and adding part-new language annotations. %EE: check meaning retained
    \item Automatically annotated datasets. This type of data construction involves minimal human participation and relies on various multimodal models for filtering and annotating image data.
\end{itemize}

In this section, we will delve into the construction methods and applicable task domains of existing remote sensing VLM datasets, offering insights into how these datasets support extensive applications and in-depth research within the field. The statistical information of current existing datasets are summarized in Table~\ref{tab:datasets}.

\subsection{Manual Datasets}
%手动制作-数量少质量高
Manually curated datasets, though limited in size, are typically of high quality and are specifically designed for tasks. There are three high-quality datasets made by hand, which are meticulously crafted to address specific challenges within the remote sensing domain. 

HallusionBench~\cite{hallusionbench}, the first benchmark designed to examine visual illusion and knowledge hallucination of large vision--language models (LVLMs) and analyze the potential failure modes based on each hand-crafted example pair, consists of 455 visual--question control pairs, including 346 different figures and a total of 1129 questions on diverse topics.
RSICap's~\cite{rsgpt} RSIEval %EE: check meaning retained
comprises five expert-annotated captions with rich and high-quality information; thus, the model achieves satisfactory performance with fine-tuning. 
CRSVQA~\cite{Multistep-VQA} employs annotators from different backgrounds, including geography, surveying, and mapping, to write complex questions that should reason the
relationship between the objects or a concise question related to the image content.
This can effectively prevent the problems of data redundancy and language bias in existing RSVQA datasets~\cite{RSVQA}.

\begin{table}[ht]
\caption{The Statistical Information of Datasets. K=Thousands, M=Milion, SC=Scene Classification, VQA=Visual Question Answer, VG=Visual Grounding, RRSIS=Referring Remote Sensing Image Segmentation, IC=Image Captioning. \label{tab:datasets}}
  \begin{tabularx}{\textwidth}{
            >{\centering\arraybackslash}m{1.8cm}
            >{\raggedright\arraybackslash}m{2.8cm}
            >{\centering\arraybackslash}m{0.4cm}
            >{\raggedright\arraybackslash}X
            >{\centering\arraybackslash}m{1.3cm}
            >{\centering\arraybackslash}m{1.1cm}
            >{\centering\arraybackslash}m{1.5cm}
            }
			\toprule
			\textbf{Method}	& \textbf{Dataset} &\textbf{Year} &\textbf{Source} &\textbf{Image} &\textbf{Image Size}  &\textbf{Task}\\
			\midrule
\multirow{3}{*}{\shortstack{Manual \\ Datasets}}  
                            &HallusionBench~\cite{hallusionbench}& 2023& - & 346& - & VQA\\
                            &RSICap~\cite{rsgpt} & 2023&DOTA & 2585& 512&IC\\
                            &CRSVQA~\cite{Multistep-VQA} &2023 & AID & 4639 & 600 &VQA\\
                            \midrule
                            
\multirow{12}{*}{\shortstack{Combining \\ Datasets}}                   
                            &SATIN~\cite{classficationSI-satin}&2023& Million-AID, WHU-RS19, SAT-4, AID \textsuperscript{1}& $\approx$775 K & - & SC\\
                            &GeoPile~\cite{Towards-geospatial-foundation-models} &2023& NAIP, RSD46-WHU, MLRSNet, RESISC45, PatternNet & 600 K & - & -\\
                            &SatlasPretrain~\cite{bastani2023satlaspretrain} & 2023&UCM, BigEarthNet, AID, Million-AID, RESISC45, FMoW, DOTA, iSAID & 856 K & 512 & - \\
                            %&CRSVQA~\cite{Multistep-VQA} & AID & 4639 & 600 &VQA\\
                            &RSVGD~\cite{rsvg}& 2023& DIOR & 17,402 & 800 &VG\\
                            &RefsegRS~\cite{rrsis} & 2024& SkyScapes & 4420  & 512&RRSIS\\
                            %&RSICap~\cite{rsgpt} &DOTA & 2585& 512&IC\\
                            &SkyEye-968 K~\cite{skyeyegpt} & 2024& RSICD, RSITMD, RSIVQA, RSVG \textsuperscript{1} &968 K& -&-\\
                            &MMRS-1M~\cite{earthgpt}&2024 & AID, RSIVQA, Syndney-Captions \textsuperscript{1} & 1 M &-& -\\
                            &RSSA~\cite{h2rsvlm}&2024 & DOTA-v2, FAIR1M & 44 K & 512 & -\\
                            &FineGrip~\cite{Panoptic-Perception}&2024 & MAR20 & 2649 & - & -\\
                            &RRSIS-D~\cite{RRSIS-D}&2024& - & 17,402  & 800 &RRSIS\\
                            &RingMo~\cite{ringmo}&2022 & Gaofen, GeoEye, WorldView, QuickBird \textsuperscript{1} & $\approx$2 M  & 448 &-\\
                            &GRAFT~\cite{RemoteSV}&2023 & NAIP, Sentinel-2 & -  & -& -\\
                            &SkySense~\cite{skysense}&2024& WorldView, Sentinel-1, Sentinel-2 & 21.5 M  & - & -\\
                            &EarthVQA~\cite{earthvqa}&2024& LoveDA, WorldView & 6000  & - & VQA\\
                            &GeoSense~\cite{Generative-ConvNet-FM}&2024& Sentinel-2, Gaofen, Landsat, QuickBird & $\approx$9 M & 224 &-\\
                            \midrule
\multirow{10}{*}{\shortstack{Automatically \\ Annotated \\Datasets}}                        
                            &RS5M~\cite{rs5mgeorscliplargescale}& 2024&LAION2B-en, LAION400M,  LAIONCOCO \textsuperscript{1}& 5M & - & -\\
                            &SkyScript~\cite{skyscript}&2024& Google Earth Engine, OpenStreetMap & 2.6 M & - & -\\
                            &LHRS-Align~\cite{Lhrs-bot} &2024& Google Earth Engine, OpenStreetMap & 1.15 M & - & -\\
                            &GeoChat~\cite{geochat} & 2024& SAMRS, NWPU-RESISC-45, LRBEN, Floodnet & 318 K & - & -\\
                            &GeoReasoner~\cite{georeasoner} &2024 & Google Street View, OpenStreetMap & 70 K+ &- & StreetView\\
                            &HqDC-1.4 M~\cite{h2rsvlm}& 2024& Million-AID, CrowdAI, fMoW, CVUSA, CVACT, LoveDA & $\approx$1.4 M &512 & - \\
                            &ChatEarthNet~\cite{chatearthnet} & 2024&  Sentinel-2, WorldCover & 163,488 & 256 & -\\
                            &VRSBench~\cite{vrsbench} &2024& DOTA-v2, DIOR & 29,614 & 512 & -\\
                            &FIT-RS~\cite{skysensegpt}&2024 &  STAR & 1800.8 K & 512 &-\\
			\bottomrule
		\end{tabularx}
	%\end{adjustwidth}
	\noindent{\footnotesize{\textsuperscript{1} Only a subset of common datasets are enumerated.}}
\end{table}

However, manual datasets are time-consuming and expensive to create, requiring domain experts to annotate data. Even with expert annotators, inconsistencies and errors may arise, affecting dataset quality. Additionally, the limited capacity for manual annotation results in smaller datasets, which can hinder model generalization.
To ensure consistency, clear annotation guidelines and multiple annotators for cross-validation are essential. Regular audits by experts can maintain quality. Using semi-automated tools can improve efficiency, helping manage large-scale manual annotation projects.

\subsection{Combining Datasets}
As shown in Table~\ref{tab:datasets}, some datasets are built from scratch, whose sources are usually Sentinel-2, Gaofen, Google Earth Engine and so on. 
But most datasets are created by merging existing remote sensing datasets, which facilitates multi-task model creation by integrating various data types from traditional RS domain-specific datasets,  and greatly eases labor compared to manual production and starting from scratch. %EE: check meaning retained
The most utilized datasets are AID~\cite{AID}, NWPU-RESISC45~\cite{NWPU}, UCM~\cite{UCM} for Scene Classification, FAIR1M~\cite{FAIR1M}, DIOR~\cite{DIOR} for object detection, iSAID~\cite{iSAID} for semantic segmentation,  and LEVIR-CD~\cite{CDdataset} for Change Detection; more details about those datasets can be found in \ref{sec:Capabilities}.
Additionally, the well-known benchmark dataset Million-AID~\cite{million-aid} is frequently included.

Combining datasets from different sources often leads to issues such as inconsistent formats, resolutions, and annotations, requiring extensive preprocessing. Misalignment of annotations can occur, and merging datasets from different contexts (e.g., regions or timeframes) may introduce gaps in data, reducing generalization.
When combining datasets, it is crucial to standardize the data formats, resolutions, and annotation styles across the datasets to ensure consistency. Preprocessing steps such as data harmonization are necessary to align different datasets and eliminate inconsistencies. Careful curation ensures balanced representation across different data sources. Filling annotation gaps and validating the combined dataset help improve consistency and robustness.

\subsection{Automatically Annoteted Datasets}
The explosive development of VLMs and LLMs has provided efficient methods for constructing large-scale remote sensing vision--language datasets. This category, which is rapidly becoming the mainstream, leverages the power of VLMs and LLMs to construct large-scale, high-quality datasets. 

RS5M~\cite{rs5mgeorscliplargescale} employs the general multimodal model BLIP2~\cite{blip-2} to generate image captions and then uses CLIP~\cite{clip} to select the top five highest-scoring captions. 
SkyScript~\cite{skyscript} matches Google Images with the OpenStreetMap (OSM) database and selects relevant image attribute values from OSM to concatenate them with CLIP~\cite{clip} as image captions.
LHRS-Align~\cite{Lhrs-bot} uses a similar approach to extract multiple attribute values from OSM and then summarizes them using Vicuna-13B~\cite{Vicuna} to produce long and fluid captions.
GeoChat~\cite{geochat} specifically provide system instructions as prompts that ask Vicuna~\cite{Vicuna} to generate multi-round question and answer pairs.
GeoReasoner~\cite{georeasoner} obtains image--text data pairs for geolocation from the communities of the two games, utilizes BERT~\cite{BERT} to clean and filter text that lacks geolocation-specific information, and then uses CLIP~\cite{clip} to align the image and text to construct a highly locatable Street View dataset.
HqDC-1.4~M~\cite{h2rsvlm} utilizes the “gemini-1.0-pro-vision''~\cite{gemini} to generate descriptions for images from multiple public RS datasets, thereby obtaining a dataset of image--text pairs to serve as the pre-training data for RSVLMs.
ChatEarthNet~\cite{chatearthnet} carefully designs prompts that embed semantic information from the land cover maps to make ChatGPT-3.5 and ChatGPT-4v achieve the best results on satellite imagery for the caption generation.
VRSBench~\cite{vrsbench} did a similar job and spent thousands of hours manually verifying the generated annotations.
FIT-RS~\cite{skysensegpt} used the efficient TinyLLaVA-3.1B~\cite{tinyllava} to swiftly generate concise background descriptions for numerous remote sensing imageries (RSIs), then combined them as the prompt to obtain detailed and fluent sentences using GPT-4/GPT-3.5.
These are examples which utilize advanced models like BLIP2\cite{blip-2} and CLIP~\cite{clip} to generate detailed image--text pairs, supporting complex applications such as disaster monitoring, urban planning, and environmental analysis. %EE: check meaning retained

Automatically generated annotations can also introduce errors, as models may misinterpret features or be biased due to limited training data. The lack of human oversight means that errors may persist, resulting in lower-quality annotations, especially for complex or sparse data.
To mitigate errors in automatically annotated datasets, models should be carefully evaluated and fine-tuned on high-quality ground-truth data. Implementing hybrid annotation strategies, where human oversight is incorporated into the process, can help identify and correct errors that may go unnoticed in fully automated processes. Data augmentation techniques can be used to enhance the dataset and address limitations in model performance, especially in cases of sparse or ambiguous data. Regular validation of automatic annotations against expert-annotated data can also help ensure the quality of the dataset. Additionally, models should be continually retrained and updated to address evolving challenges in remote sensing and improve annotation accuracy.

\subsection{Summary}
The existing three types of datasets each have their own advantages and disadvantages, as detailed below:
\begin{itemize}
    \item Manually annotated datasets generally have higher quality and are most closely aligned with specific problems. But these datasets, typically consisting of only a few hundred or thousand images, are limited in scale compared to combined datasets and VLM datasets, which can reach tens of millions of images. The size of a dataset is crucial for model training, as larger and richer datasets yield more accurate and versatile models. Manually annotated datasets can be used to fine-tune large models, but they struggle to support training large models from scratch.    
    \item Combined datasets. By merging existing datasets, large-scale datasets can be constructed at a low cost. This type of data can easily reach millions of entries. However, datasets created in this manner typically have lower annotation quality compared to manually annotated datasets and may not fully align with specific tasks. Nonetheless, they can be utilized for model pre-training.
    
    % Large-scale dataset construction involves merging various types of datasets for multiple tasks, necessitating complex automatic or semi-automatic annotation methods to ensure versatility and efficiency, along with manual sampling detection to maintain quality.
    \item Automatically annotated datasets. Mainstream VLM dataset production methods leverage state-of-the-art LLMs and VLMs to generate high-quality captions or annotations for large-scale remote sensing images. This approach produces flexible image--text pairs, including long sentences, comments, and local descriptions, without requiring human supervision, thus meeting the needs of various tasks. Researchers may need to craft prompts and commands to guide LLMs and VLMs in generating the desired texts.     
\end{itemize}

Overall, automatically annotated datasets leverage the advantages of pre-trained large models, resulting in datasets that are both quantitatively and qualitatively substantial. As a result, this type of dataset has become quite mainstream in the field of remote sensing.

\section{Capabilities}
\label{sec:Capabilities}

\begin{figure}[t]
	\centering
	\includegraphics[width=0.6\textwidth]{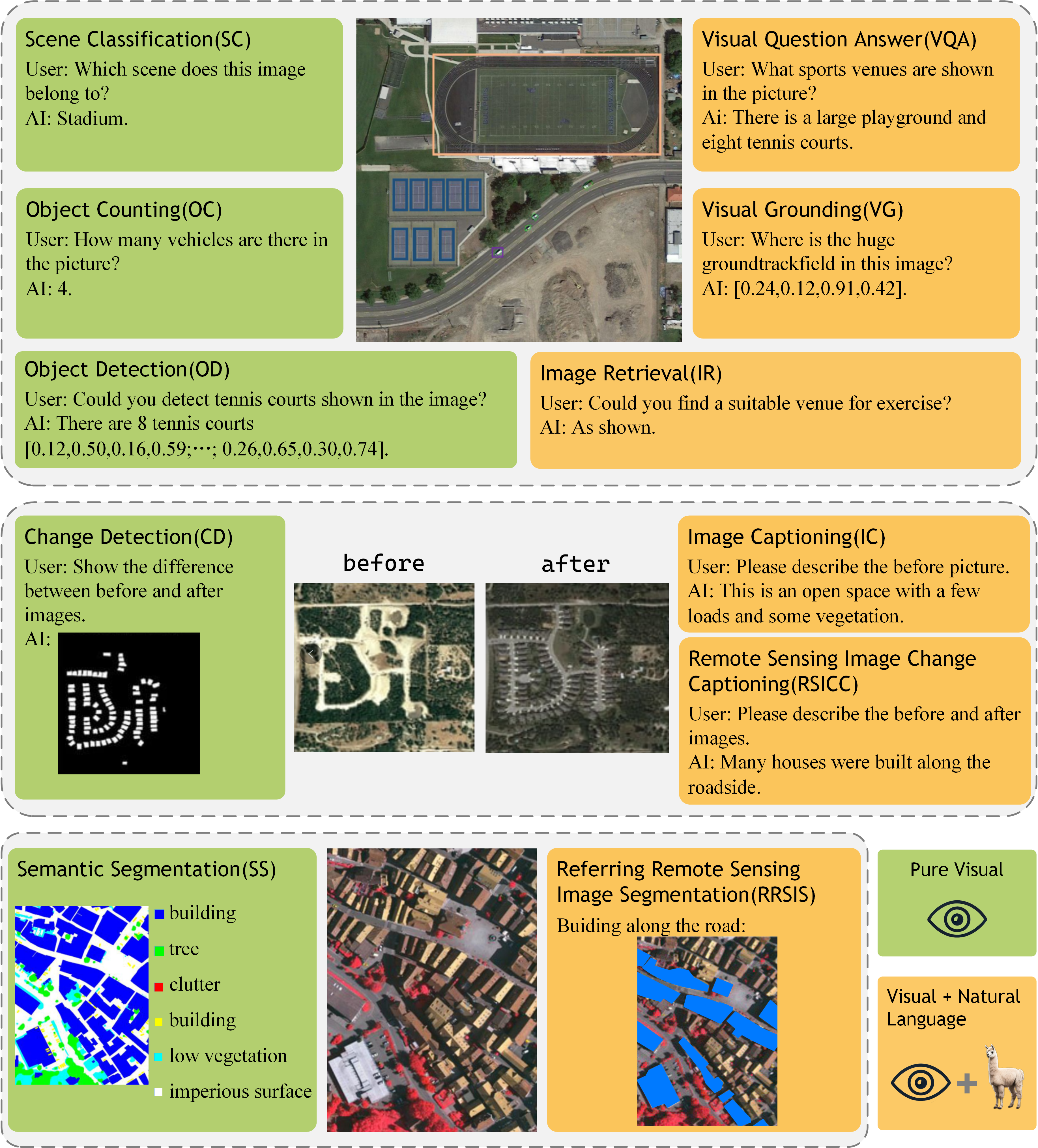}
    \captionsetup{justification=centering}
	\caption{Examples of tasks handled by VLMs in remote sensing field}
	\label{fig:rs_exmaples}
 \end{figure}

%多任务趋势要介绍
In various tasks within the field of remote sensing, VLMs have demonstrated powerful capabilities. These capabilities can be broadly classified into two categories: pure visual tasks and vision--language tasks.  
Pure visual tasks focus on analyzing remote sensing images to extract meaningful information about the Earth's surface. Vision--language tasks, which combine natural language processing with visual data analysis, have opened new avenues for remote sensing applications.
The zero-shot and few-shot tasks are SC, OD, SS and IR, which are described in more detail below.
Figure~\ref{fig:rs_exmaples} illustrates each capability with a simple example, particularly highlighting two emerging tasks along with their source tasks.

\subsection{Pure visual}

\begin{itemize}
%基础任务
%基础任务近年来few-shot和zero-shot
\item   Scene Classification (SC): %MDPI: 1. Please confirm if the italics is unnecessary and can be removed. 2. Please check all italics in the main text. 3. For reducing space, Figure 8 across heading 4.1, please confirm
 RS Scene Classification entails the classification of satellite or aerial images into distinct land-cover or land-use categories with the objective of deriving valuable insights about the Earth’s surface. It offers valuable information regarding the spatial distribution and temporal changes in various land-cover categories, such as forests, agricultural fields, water bodies, urban areas, and natural landscapes. The three most commonly used datasets in this field are AID~\cite{AID}, NWPU-RESISC45~\cite{NWPU} and UCM~\cite{UCM}, which contain 10000, 31500, and 2100 images, and 30, 45, and 21 categories, respectively. 
\item   Object Detection (OD): RSOD aims to determine whether or not objects of interest exist in a given RSI and  return the category and position of each predicted object. In this domain, the most widely used datasets are DOTA~\cite{DOTA} and DIOR~\cite{DIOR}. DOTA~\cite{DOTA} contains 2806 aerial images which are annotated by experts in aerial image interpretation, with respect to 15 common object categories. DIOR~\cite{DIOR} contains 23463 images and 192,472 instances, covering 20 object classes, with much more detail than DOTA~\cite{DOTA}.
\item   Semantic Segmentation (SS): Semantic segmentation is a computer vision task in which the goal is to categorize each pixel in an image into a class or object. For RS images, it plays an irreplaceable role in disaster assessment, crop yield estimation, and land change monitoring. The most widely used datasets are ISPRS Vaihingen and Potsdam, which contain 33 true orthophoto (TOP) images and 38 image tiles with a spatial resolution of 5 cm, respectively. Another widely used dataset iSAID~\cite{iSAID} contains 2806~aerial images, which were mainly collected
from Google Earth.
\item   Change Detection (CD): Change Detection in remote sensing refers to the process of identifying differences in the structure of objects and phenomena on Earth surface by analyzing two or more images taken at different times. It plays an important role in urban expansion, deforestation, and damage assessment. LEVIR-CD~\cite{CDdataset}, AICD~\cite{AICD} and the Google Data Set focus on changes in buildings. Other types of changes such as roads and groundwork are commonly included in CD datasets.

\item   Object Counting (OC): RSOC aims to automatically estimate the number of object instances in an RSI. It plays an important role in many areas, such as urban planning, disaster response and assessment, environment control, and mapping. RemoteCount~\cite{remoteclip} is a manual dataset for object counting in remote sensing imagery consisting of 947~image--text pairs and 13 categories, which are mainly selected from the validation set of the DOTA dataset. %也是zero-shot
\end{itemize}

\subsection{Combine with natural language}
\begin{itemize}
%复杂任务--和natural language更相关的
\item   Image Retrieval (IR): IR from RS aims to retrieve  RS images  of interest from massive RS image repositories. With the launch of more and more Earth observation satellites and the emergence of large-scale remote sensing datasets, RS image retrieval can prepare auxiliary data or narrow the search space for a large number of RS image processing tasks.
\item   Visual Question Answering (VQA): VQA is a task that seeks to provide answers to free-form and open-ended questions about a given image. As the questions can be unconstrained, a VQA model applied to remote sensing data could serve as a generic solution to classical problems but also very specific tasks involving relations between objects of different nature~\cite{survey2}. VQA was first proposed by Anto et~al.~\cite{VQA} in 2015, and then  applied to the RS domain by RSVQA~\cite{RSVQA} in 2020. This work contributed an RSVQA framework and two VQA datasets, RSVQA-LR and RSVQA-HR, which are widely utilized in further works. %EE: check meaning retained
\item   Image Captioning (IC): IC aims to generate natural language descriptions that summarize the content of an image. It requires representing the semantic relationships among objects and generating an exact and descriptive sentence, so it is more complex and challenging than image detection, classification, and segmentation tasks. In this domain, the commonly used datasets are Syndney-Captions~\cite{captiondatasets}, RSICD~\cite{RSICD}, NWPU-Captions~\cite{NWPUCaptions}, and UCM-Captions~\cite{captiondatasets}. Recently, CapERA~\cite{capera} provided UAV video with diverse textual descriptions to advance visual–language-understanding tasks.
\item   Visual Grounding (VG): RSVG aims to localize specific objects in remote sensing images with the guidance of natural language. %EE: check meaning retained
It was first introduced in~\cite{first-rsvg} in 2022. In 2023, Yang~\cite{rsvg} et~al. not only built the new large-scale benchmark of RSVG based on detection in the DIOR dataset, termed RSVGD, but designed a novel transformer-based MGVLF module to address the problems of scale variation and cluttered background in RS images.
\item   Remote Sensing Image Change Captioning (RSICC): A new task aiming to generate human-like language descriptions for the land-cover changes in multitemporal RS images. It is a combination of IC and CD tasks, offering important application prospects in damage assessment, environmental protection, and land planning. Chenyang et~al.~\cite{RSICC} first introduced the CC task into the RS domain in 2022 and proposed a large-scale dataset LEVIR-CC, containing 10,077 pairs of bitemporal RS images and 50,385 sentences describing the differences between the images. In 2023, they proposed a pure Transformer-based model~\cite{a-decoupling} to further improve the performance of the RSICC task.
\item   Referring Remote Sensing Image Segmentation (RRSIS): RRSIS  provides a pixel-level mask of desired objects based on the content of given remote sensing images and natural language expressions. It was first proposed by Zhenghang et~al.~\cite{rrsis} in 2024, who created a new dataset, called RefSegRS, for this task. In the same year, Sihan et~al.~\cite{RRSIS-D} curated an expansive dataset comprising 17,402 image--caption--mask triplets, called RRSIS-D, and proposed a series of models to meet the ubiquitous rotational phenomena in RRSIS.
\end{itemize}

\begin{figure}[t]
\centerline{\includegraphics[width=13.5 cm]{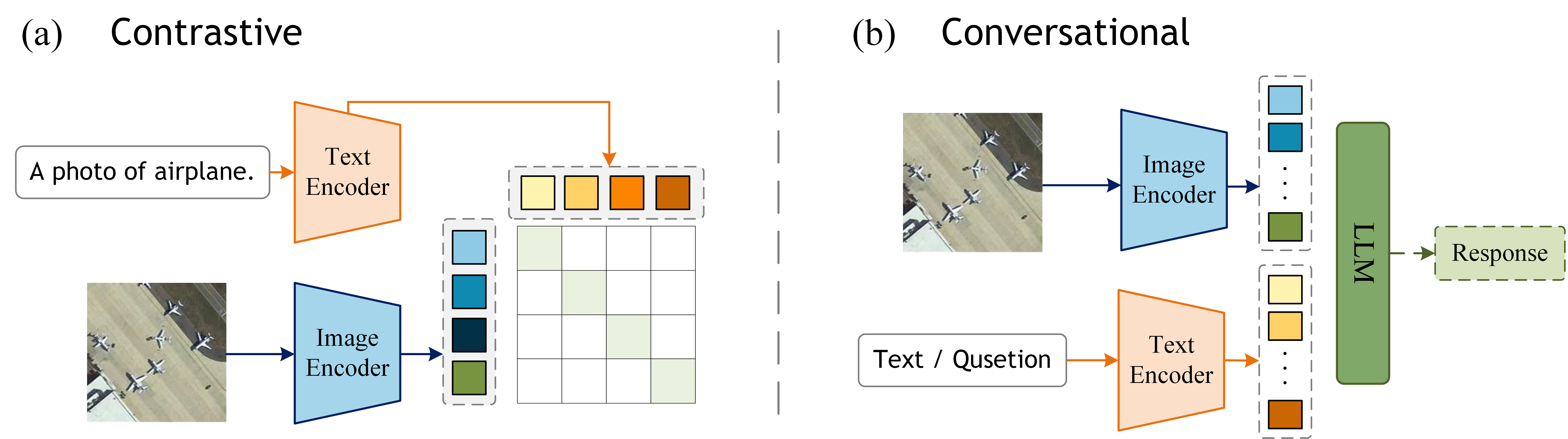}}
%\captionsetup{justification=centering}
\caption{Fundamental architecture of contrastive and conversational Vision--Language Models. 
\label{fig:architecture}}
\end{figure}  

The invention of VLMs has greatly changed the remote sensing field, which mainly focuses on geospatial analysis. With the development and usage of various sensors, conventional 
vision models cannot handle such complex data types, while VLMs could  enhance the accuracy and efficiency of remote sensing analysis. Specifically, the combination of visual and linguistic
data within VLMs allows more subtle and sophisticated interpretations of complex remote sensing datasets. By utilizing their multimodal capabilities, VLMs
have shown significant progress across various tasks within remote sensing, including object detection, geophysical classification, and scene understanding.
Meanwhile, they provide more
accurate and smart solutions for practical applications such as disaster monitoring,
urban planning, and agricultural management.

\section{Recent Advances}
\label{sec:Enhancement Techniques}

Currently, there are two main types of models that integrate vision and language: one is based on contrastive learning structures, such as the CLIP series, and the other is based on large language models that fuse visual features, like LLaVA, as shown in Figure~\ref{fig:architecture}. Accordingly, recent improvements %EE: check meaning retained
can also be categorized into two major types: enhancements within the contrastive learning framework and improvements within the conversational framework.

\subsection{Advancements in Contrastive VLMs}

Contrastive VLMs primarily consist of an image encoder and a text encoder, with the core challenge being the alignment of features extracted by both encoders. As shown in Table~\ref{tab:contrastive}, the image encoder, utilizing a CNN or ViT, encodes images into feature vectors. Concurrently, the text encoder processes textual descriptions into corresponding feature vectors. The model is trained to maximize the similarity between positive image--text pairs while minimizing the similarity between negative pairs. CLIP is a pioneering model in vision--language pre-training, employing InfoNCE loss to align text and image features in a shared space. Currently, in the field of remote sensing, contrastive VLMs focus on transferring the image--text alignment capabilities established in RGB images and text to remote sensing images and text, which are then utilized to address various downstream tasks, such as Change Detection and remote sensing Image Captioning.

\begin{table}[t] 
\caption{Architecture of contrastive VLMs. SC = Scene Classification; VG = Visual Grounding; \mbox{IC = Image} Captioning; CD = Change Detection; IR = Image Retrieval. For each model, we indicate the LLM used in its best configuration as shown in the original paper.\label{tab:contrastive}}
\renewcommand{\arraystretch}{1.1}
\begin{tabularx}{\textwidth}{
 >{\centering\arraybackslash}m{3cm}
  >{\centering\arraybackslash}m{1cm}
   >{\centering\arraybackslash}m{3cm}
    >{\centering\arraybackslash}X
     >{\centering\arraybackslash}X
}
\toprule
\textbf{Model}	& \textbf{Year} & \textbf{Image Encoder}	& \textbf{Text Encoder } & \textbf{Capability} \\
\midrule

%\rowcolor{cyan!10}
RemoteCLIP~\cite{remoteclip}  &2024          & ViT-B (CLIP)                                               & Transformer                                                   & SC, IC, IR, OC              \\
%\rowcolor{cyan!10}
CRSR~\cite{cross-modal}    &2024             & ViT-B (CLIP)                                                &       -                                                        & IC                       \\
%\rowcolor{cyan!10}
ProGEO~\cite{progeo}     &2024               & ViT-B (CLIP)                                                & BERT                                                          & VG                       \\
%\rowcolor{cyan!10}
GRAFT~\cite{RemoteSV}   &2023                & ViT-B (CLIP)                                                     &       -                                                        & SC, VQA, IR               \\
%\rowcolor{cyan!10}
GeoRSCLIP~\cite{rs5mgeorscliplargescale} & 2024& ViT-H (CLIP)                                               &      -                                                         & SC, IR                    \\
%\rowcolor{cyan!10}
ChangeCLIP~\cite{changeclip} &2024           & CLIP                                                       &        -                                                       & CD                       \\
%\rowcolor{cyan!10}
APPLeNet~\cite{applenet} &2023               & CLIP                                                    &     -                                                          & SC, IC                    \\
%\rowcolor{cyan!10}
MGVLF~\cite{rsvg}   &  2023                  & ResNet-50                                                  & BERT                                                          & VG                       \\
\bottomrule
\end{tabularx}
\end{table}

To address the issue of requiring additional annotated data and fine-tuning when building foundational models using self-supervised learning and masked image modeling methods in the remote sensing domain, Liu et~al.~\cite{remoteclip} proposed the first vision--language foundational model for remote sensing, RemoteCLIP. This model is designed to learn robust visual features with rich semantics and aligned text embeddings for seamless downstream applications. They created a combined image--caption dataset and fine-tuned the CLIP model on this newly constructed dataset through continual learning, thereby tackling the scarcity of pre-trained data. RemoteCLIP exhibits powerful multi-task capabilities and can be applied to a range of downstream tasks, including zero-shot image classification, linear probing, k-NN classification, few-shot classification, image-text retrieval, and object counting.
%------David-----%
Similar to~\cite{remoteclip}, GeoRSCLIP builds a remote sensing dataset consisting of 5 million image--text pairs by filtering existing image--text pair datasets and generating captions for image-only datasets. It then fine-tunes CLIP on this proposed dataset~\cite{rs5mgeorscliplargescale}. CRSR~\cite{cross-modal} incorporates a Transformer Mapper network, utilizing an attention mechanism and learnable semantic predictive queries to provide a comprehensive representation of image features. This approach effectively analyzes and understands more semantic details present in RSI. ProGEO proposes adding more descriptions to a geographic image-based pre-trained CLIP, using these text descriptions to enhance the image encoder by aligning features from the image encoder with the features of the descriptions~\cite{progeo}. This approach enables the image encoder to generate features with more details, leading to improvements in visual geo-localization. CRAFT proposes training a satellite image model to align its embedding with the CLIP embedding of co-located internet images, which sidestep the need for textual annotations for training remote sensing VLMs~\cite{RemoteSV}. ChangeCLIP~\cite{changeclip} is the first to apply a multimodal vision--language scheme to Remote Sensing Change Detection tasks.  It introduces a Differentiable Feature Calculation (DFC) layer, enabling the network to focus on learning the changed features of bitemporal remote sensing images rather than semantic categories. APPLeNet designs an injection module to fuse image features from different levels and interact with text to generate better visual tokens~\cite{applenet}.

In addition, S-CLIP~\cite{sclip} and RS-CLIP~\cite{rs-clip} enhance the performance of CLIP in remote sensing by improving the pseudo-label strategy. The former divides pseudo-labels into title-level and keyword-level categories, combining their advantages to ensure the model's generalization ability. RS-CLIP automatically generates pseudo-labels from unlabeled data and employs a curriculum learning strategy to gradually select more pseudo-samples for model fine-tuning.

\begin{figure}
	\centering
	\includegraphics[width=0.8\textwidth]{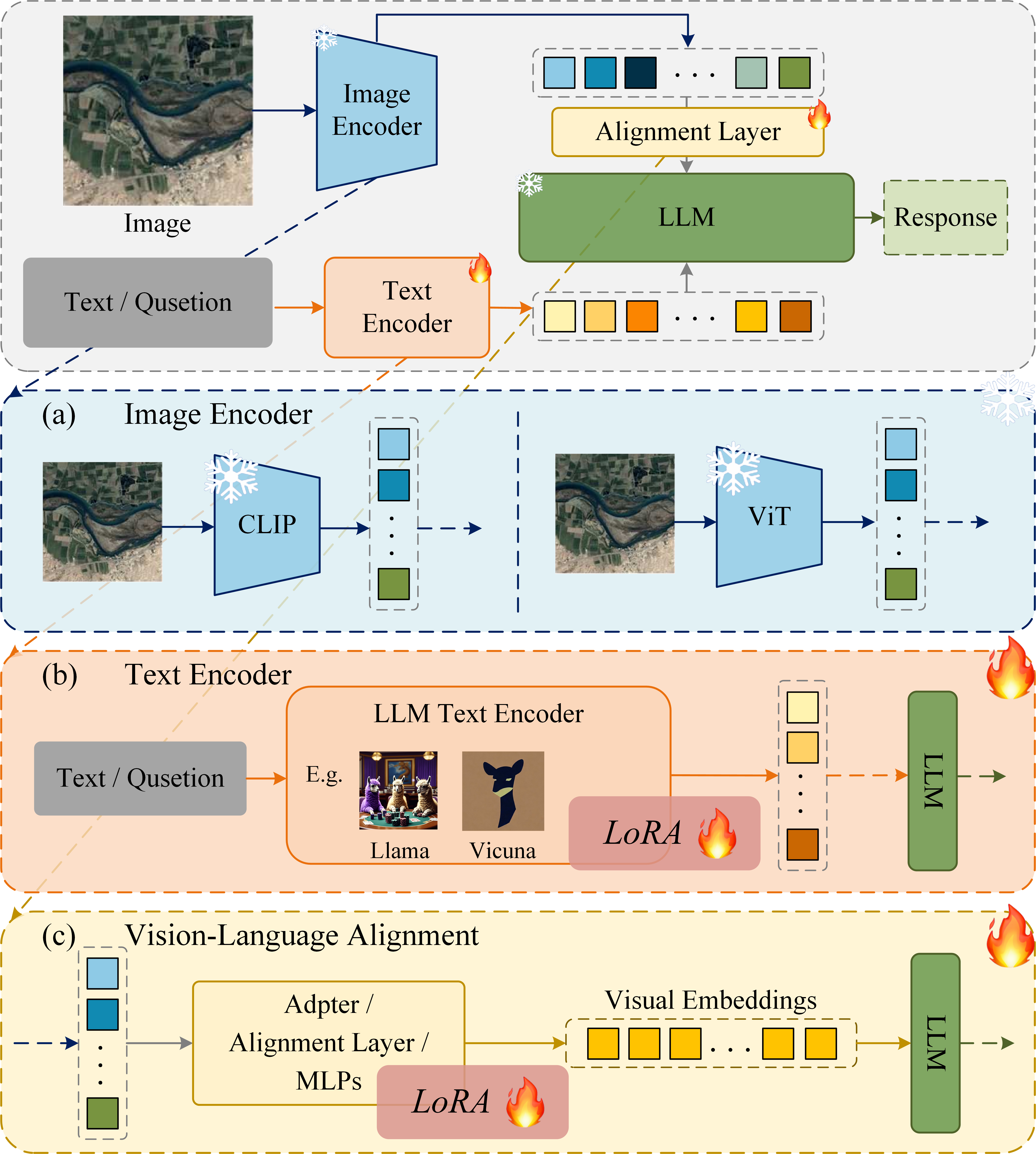}
    \captionsetup{justification=centering}
	\caption{
		Enhancement Techniques
	}
	\label{fig:enhanctech}
\end{figure}

\subsection{Advancements in Conversational VLMs}

Currently, conversational VLMs are particularly noteworthy for their ability to address complex remote sensing tasks by transforming intricate visual features into a format that LLMs can comprehend. By integrating advanced image encoders that adeptly capture subtle variations in remote sensing imagery with text encoders optimized for efficiently processing language instructions, these models demonstrate remarkable performance, particularly in few-shot learning scenarios. The alignment layer is pivotal in establishing connections between remote sensing imagery and LLMs. Techniques such as utilizing MLPs for feature projection and incorporating learnable query embeddings for task-specific image feature extraction have proven effective in elevating the overall performance of VLMs.

Leveraging the existing API of pre-trained LLMs, conversational VLMs extract features from image data using a visual encoder. These features are then transformed into a format that LLMs can interpret, enabling the completion of various vision--language tasks. To enhance task performance, researchers meticulously design each component, which is described in detail below. Figure~\ref{fig:enhanctech} clearly shows the model details of conversational VLMs.

Image Encoder: Table~\ref{tab:conversational} presents the visual encoders commonly used in current VLM implementations. 
Most studies utilize the standard CLIP image encoder~\cite{clip}, but RSGPT~\cite{rsgpt} and SkyEyeGPT~\cite{skyeyegpt} have explored other variants, such as EVA-CLIP encoder~\cite{eva-clip}, which is trained with improved training techniques. 

Additionally, RS-LLaVa~\cite{RS-llava} adapts the LLaVA framework for the remote sensing domain by using a domain-specific image encoder and training it on specialized RS datasets. Through instruction tuning and precise visual--language alignment, RS-LLaVa gains a deeper understanding of geospatial semantics, allowing it to provide contextually relevant, detail-oriented responses tailored to the complexities of remote sensing imagery. 
EarthGPT~\cite{earthgpt} employs a Vision Transformer as its image encoder to extract multi-layered features that capture subtle differences in RS images. It then integrates a CNN backbone to incorporate multi-scale local details into the visual representation, delivering more comprehensive analytical capabilities.
GeoChat~\cite{geochat} builds on the CLIP-ViT(L-14) backbone, interpolating positional encodings at higher resolutions (512 × 512) to better discern fine-grained details. This approach enhances its capacity to identify small objects and subtle characteristics in remote sensing imagery.

\begin{table}[ht] 
\caption{Architecture of conversational VLMs. $\ast$: frozen; $\blacktriangle$: fine-tuning; SC = Scene Classification; VQA = Visual Question Answering; VG = Visual Grounding; IC = Image Captioning; OD = object detection; IR = Image Retrieval; RSICC = Remote Sensing Image Change Captioning. For each model, we indicate the LLM used in its best configuration as shown in the original paper.}
\label{tab:conversational}
\begin{tabularx}{\textwidth}{
>{\centering\arraybackslash}m{3.5cm}
  >{\centering\arraybackslash}m{1cm}
   >{\centering\arraybackslash}X
    >{\centering\arraybackslash}X
     >{\centering\arraybackslash}m{4.5cm}
}
\toprule
\textbf{Model}&\textbf{Year}& \textbf{Image Encoder}	& \textbf{Text Encoder } & \textbf{Capability} \\
\midrule
RS-LLaVA~\cite{RS-llava} &   2024            & ViT-L(CLIP)~$^\ast$                                               & Vicuna-v1.5-13B~$^\ast$                                             & IC, VQA                   \\
H2RSVLM~\cite{h2rsvlm}  &2024                & ViT-L (CLIP)~$^\ast$                       & Vicuna-v1.5-7B~$^\ast$                        & SC, VQA, VG                \\
SkySenseGPT~\cite{skysensegpt} & 2024        & ViT-L (CLIP)~$^\ast$                                              & Vicuna-v1.5-7B~$^\blacktriangle$                                & SC, IC, VQA, OD             \\
GeoChat~\cite{geochat}  &2024                & ViT-L (CLIP)~$^\ast$                                              & Vicuna-v1.5-7B~$^\blacktriangle$                                & SC, VQA, VG\par Region Caption \\
RSGPT~\cite{rsgpt}     &2023                 & ViT-G (EVA)~$^\ast$                                               & Vicuna-v1.5-7B~$^\ast$                                               & IC, VQA                   \\
SkyEyeGPT~\cite{skyeyegpt}  &2024            & ViT-G (EVA)~$^\ast$                                                & LLaMA2-7B~$^\ast$                                                    & IC, VQA, VG                \\
RS-CapRet~\cite{RS-CapRet}   &2024           & ViT-L (CLIP)~$^\blacktriangle$                                & LLaMA2-7B~$^\ast$                                                    & IC, IR                    \\
LHRS-Bot~\cite{Lhrs-bot}   &2024             & ViT-L (CLIP)~$^\ast$                                              & LLaMA2-7B~$^\ast$                                                    & SC, VQA, VG                \\
EarthGPT~\cite{earthgpt} &2024               & ViT-L (DINOv2)~$^\ast$ \par ConvNeXt-L (CLIP)~$^\ast$                         & LLaMA2-7B~$^\blacktriangle$                                     & SC, VQA, IC, VG, OD          \\
Liu et~al.~\cite{a-decoupling}  &2023        & ViT-B(CLIP)~$^\ast$                                                      & GPT-2~$^\ast$                                                        & RSICC                    \\
\bottomrule
\end{tabularx}
\end{table}

Text Encoder: Training large language models (LLMs) from scratch is highly costly. Therefore, conversational Vision--Language Models (VLMs) utilize pre-trained LLMs as text encoders. These LLMs process visual elements extracted by the visual encoder and converted by middleware, along with the researcher's questions or language instructions, to generate the desired results. Currently, the academic community is focusing on the LLaMA~\cite{LLaVA} and Vicuna~\cite{Vicuna} series, which serve as text encoders for most conversational LLMs. Table~\ref{tab:conversational} %MDPI: The first citation of each table should appear in numerical order, Table 3 should be after Table 2, there are no Table 2 cited before, please revise.
 provides a summary of these models. Common approaches include directly using LLMs or efficiently fine-tuning them with Low-Rank Adaptation (LoRA)~\cite{LoRA}. 
 Liu et~al.~\cite{a-decoupling} designed a complex image classifier and proposed a multi-prompt learning strategy to generate multiple learnable prompt embeddings as prefixes for LLMs to enhance their utilization for the RSICC task. In addition, RS-LLaVA~\cite{RS-llava} applies the LLaVA framework to remote sensing data, incorporating a Vicuna-v1.5-13B LLM along with domain-specific image encoders and specialized training corpora, enabling the model to better capture geospatial semantics. Similarly, RSGPT~\cite{rsgpt} utilizes a Vicuna-v1.5-7B LLM, showing that while large language models %EE: check meaning retained
generally yield improved performance in remote sensing tasks, considerations regarding training time and resource requirements remain central to practical deployments.

%%%
Vision--Language Alignment: In the architecture of conversational Vision--Language Models, the middle connection layer plays a crucial role. It acts as a bridge between remote sensing images and large language models, projecting image information into a space that LLMs can comprehend, enabling them to understand complex remote sensing tasks. A common approach to bridge the modal gap is to use Multi-Layer Perceptrons (MLPs), which employ a single linear layer~\cite{skyeyegpt,earthgpt} or double layers~\cite{h2rsvlm,skysensegpt,RS-llava} to project visual tokens and align feature dimensions with word embeddings. Additionally, some works focus on designing connectors with intricate structures, offering valuable insights for future research. %For instance, 

SkyEyeGPT~\cite{skyeyegpt} is introduced as a unified multimodal large language model specifically designed for remote sensing vision--language understanding, with the objective of exploring the application of VLMs in the remote sensing field. The authors carefully create a remote sensing multimodal instruction-following dataset, which includes both single-task and multi-task dialogue instructions. After manual verification, the dataset consists of a high-quality remote sensing instruction-following corpus with a total of 968 K samples. SkyEyeGPT projects remote sensing visual features into the language domain via an alignment layer, and these features, along with task-specific instructions, are fed into a remote sensing decoder based on LLM to predict answers for open-ended remote sensing tasks. SkyEyeGPT employs a two-stage fine-tuning approach to enhance instruction-following and multi-turn dialogue capabilities at different granularities. Notably, it achieves excellent performance across a variety of tasks without requiring additional encoding modules.
%------David-----%

One reason current general VLMs perform poorly in the remote sensing field is that the visual encoder of LLMs usually uses CLIP, but CLIP’s ViT lacks spatial awareness, affecting VLM’s spatial understanding ability~\cite{clip-awereness}. To address this issue, H2RSVLM~\cite{h2rsvlm} and SkySenseGPT~\cite{skysensegpt} enhance the visual encoder’s perceptual ability at the source by creating higher-quality, large-scale, and detailed image--text pairs in remote sensing datasets. EarthGPT~\cite{earthgpt} expands the input images to include SAR and infrared images. Furthermore, some research endeavors opt for designing elaborate vision--language connection structures to enhance the multimodal perception capabilities of VLM. GeoChat~\cite{geochat} constructs bounding boxes for text representations to express their spatial coordinates, enabling it to perform object grounding or accept region inputs, thus having reasoning capabilities for remote sensing images. RSGPT~\cite{rsgpt}, under the guidance of InstructBLIP~\cite{instructBLIP}, inserts an annotation-aware query transformer (Q-Former) to enhance the alignment of visual and textual features. Additionally, the annotation-aware mechanism interacts with query embeddings through self-attention layers, allowing RSGPT to extract comprehensive visual features crucial for RS vision--language tasks. LHRS-Bot~\cite{Lhrs-bot} introduces a set of learnable queries for each level of image features. These queries aim to summarize the semantic information of each level of visual features through a series of stacked cross-attention and MLP layers.

\begin{table}[t] 
\caption{Summary of other VLMs in Section \ref{section: others}.\label{tab:others}}
\begin{tabularx}{\textwidth}{
 >{\centering\arraybackslash}m{2.8cm}
  >{\centering\arraybackslash}m{0.8cm}
 >{\raggedright\arraybackslash}X
}
\toprule
\textbf{Model}&\textbf{Year}	& \textbf{Enhancement Technique} \\
\midrule
Txt2Img~\cite{txt2img} & 2023 & Employs modern Hopfield layers for hierarchical prototype learning of text and image embeddings. \\     
CPSeg~\cite{cpseg}&2024 & Utilizes a ``chain of thought'' process that leverages textual information associated with the image.  \\
SHRNet~\cite{A_Spatial_H}&2023  & Utilizes a hash-based architecture to extract multi-scale image features. \\
MGeo~\cite{mgeo}&2023 &  Introduces the geographic environment (GC) extracting multimodal correlations for accurate query--POI matching.   \\
GeoCLIP~\cite{geoclip}&2023 & Via an image-to-GPS retrieval approach by explicitly aligning image features with corresponding GPS locations.  \\
SpectralGPT~\cite{spectralgpt} &2024& Provides a foundational model for the spectral data  based on the MAE~\cite{MAE} architecture.  \\
TEMO~\cite{FSOD} &2023 &  Captures long-term dependencies from description sentences of varying lengths, generating text-modal features for each category. \\
\bottomrule
\end{tabularx}
\end{table}

\subsection{Others Models}
\label{section: others}

In addition to the aforementioned approaches, as shown in Table~\ref{tab:others}, several other advancements have been made to address domain-specific tasks in remote sensing.

Txt2Img~\cite{txt2img} employs modern Hopfield layers for hierarchical prototype learning of text and image embeddings, enabling the generated remote sensing images to better preserve the shape information of ground objects while providing more detailed texture and boundary information. CPSeg~\cite{cpseg} employs a novel ``Chain-of-Thought'' approach using language prompting to enhance semantic segmentation in remote sensing imagery, achieving superior performance, particularly in flood scenarios, through improved context-aware understanding and integration of textual descriptions. SHRNet~\cite{A_Spatial_H} utilizes a hash-based architecture to extract multi-scale image features, while QRN improves alignment efficiency and enhances the visual--spatial reasoning ability of remote sensing Visual Question Answering systems. 
MGeo~\cite{mgeo}, a multimodal geographic language model, introduces the geographic environment (GC) as a new modality to enhance query--POI (Point of Interest) matching, outperforming baselines even without user geolocation, while its GC modeling holds potential for further improvements with extensions like POI images. 
GeoCLIP~\cite{geoclip}, the first CLIP-inspired image-to-GPS retrieval approach, introduces a novel Location Encoder that models the Earth as a continuous function using positional encoding and hierarchical representations. By aligning image features with GPS locations, GeoCLIP achieves competitive geo-localization performance, even with limited training data, and enables text-based geo-localization via its CLIP backbone.
SpectralGPT~\cite{spectralgpt}, a spectral RS foundational model based on MAE~\cite{MAE}, leverages 3D token generation and multi-target reconstruction to capture spatial--spectral patterns. Trained on one million spectral images with 600 M+ parameters, it achieves 99.21\% for single-label Scene Classification on EuroSAT~\cite{EuroSAT}  %MDPI: The reference order is wrong. We revised references order to make sure that the first citation of each reference appears in numerical order, please check and confirm.
 and also demonstrates strong performance in semantic segmentation and Change Detection, advancing spectral RS big data applications.
%这个spectralgpt似乎是有多模态的尝试的 
TEMO~\cite{FSOD} introduces a text-modal knowledge extractor and a crossmodal assembly module to fuse text and visual features, reducing classification confusion in novel classes and enhancing few-shot object detection in remote sensing imagery. By leveraging a mask strategy, separation loss, and improved word embeddings to capture long-term dependencies, TEMO achieves 42.8\%, 75.1\% and 40.8\% across DIOR~\cite{DIOR}, NWPU~\cite{nwpu-VHR10}, and FAIR1M~\cite{FAIR1M} in 10-shot scenarios, respectively.

\subsection{Performance Comparison}

In this section, we conduct a comparative analysis of recent Visual--Language Models, evaluating their performance on five widely recognized datasets and common visual--language tasks including Visual Question Answering and Image Captioning. VLMs have gained significant attention in the remote sensing domain over the past two years, with researchers exploring their applications in various tasks. However, models tailored for specialized domains, such as SpectralGPT~\cite{spectralgpt} which is designed for spectral remote sensing, face challenges in cross-model comparisons due to unique datasets and tasks. Despite efforts to build versatile foundational models capable of handling diverse downstream tasks, the remote sensing field still lacks a unified benchmark for comprehensive performance evaluation. To address this gap, we focus on shared datasets and tasks to provide a consistent and meaningful comparison of VLM performance.

Table~\ref{Performance} summarizes the evaluation results for five prevalent datasets (e.g., AID~\cite{AID}, RSVGD~\cite{rsvg}), highlighting the best-performing VLMs. Due to differing implementations of VLM pre-training methods, it is essential to identify standout models. Among the contrastive approaches, RemoteCLIP~\cite{remoteclip} emerges as a top performer, particularly excelling in WHU-RS19~\cite{WHU-RS19}. Notably, GeoRSCLIP~\cite{rs5mgeorscliplargescale} demonstrates strong few-shot capabilities, achieving competitive results on NWPU-RESISC45~\cite{NWPU}. In comparison, conversational models, such as SkySenseGPT~\cite{skysensegpt} and SkyEyeGPT~\cite{skyeyegpt}, consistently achieve superior performance in multiple datasets, with SkySenseGPT leading in AID~\cite{AID} and WHU-RS19~\cite{WHU-RS19}. Unlike contrastive models, conversational methods also support Visual Grounding tasks, highlighting their broader applicability. 

\begin{table}[t]
\caption{Performance of VLM pre-training methods on common remote sensing datasets; bold indicates the best performance. $\ast$: few-shot prediction.\label{Performance}}
\begin{tabularx}{\textwidth}{
    >{\centering\arraybackslash}m{2cm}
    >{\raggedright\arraybackslash}m{2.5cm}
    >{\centering\arraybackslash}m{1.3cm}
    >{\centering\arraybackslash}m{1.8cm}
    >{\centering\arraybackslash}m{2.2cm}
    >{\centering\arraybackslash}m{1.8cm}
    >{\centering\arraybackslash}X
}
\toprule
\textbf{Category}    & \textbf{Model} 	& \textbf{AID}~\cite{AID} & \textbf{RSVGD}~\cite{rsvg} & \textbf{NWPU-RESISC45}~\cite{NWPU} &\textbf{WHU-RS19}~\cite{WHU-RS19} & \textbf{EuroSAT}~\cite{EuroSAT} \\
\midrule
\multirow{4}{*}{Contrastive}    &RemoteCLIP~\cite{remoteclip}             & 91.30          &  -              & 79.84           & 96.12          & 60.21          \\
                                &GeoRSCLIP~\cite{rs5mgeorscliplargescale} & 76.33 $^\ast$      & -      & 73.83 $^\ast$           &   -             & \textbf{67.47} \\
                                &APPLeNet~\cite{applenet}                 &  -              &  -              & 72.73 $^\ast$         &  -              &  -              \\
                                &GRAFT~\cite{RemoteSV}                    &  -              &  -              &  -              & -               & 63.76          \\
\midrule
\multirow{6}{*}{Conversational} &SkySenseGPT~\cite{skysensegpt}           & \textbf{92.25}  &  -                      &  -              & \textbf{97.02} & -               \\
                                &LHRS-Bot~\cite{Lhrs-bot}                  & 91.26          & 88.10         &   -             & 93.17          & 51.40          \\
                                &H2RSVLM~\cite{h2rsvlm}                   & 89.33          & 48.04 $^\ast$         &   -             & 97.00          & -               \\
                                &GeoChat~\cite{geochat}                   & 72.03 $^\ast$      &  -                    &  -              &   -             & -               \\
                                &SkyEyeGPT~\cite{skyeyegpt}               &   -              & \textbf{88.59}&  -              &   -             & -               \\
                                &EarthGPT~\cite{earthgpt}                 & -               & 81.54          & \textbf{93.84}          &   -             &  -              \\
\midrule
\multirow{1}{*}{Others}  &RSVG~\cite{rsvg}                     &  -               & 78.41         &  -              &  -              &   -             \\
                        %SkySense~\cite{skysense}                 & -               & \textbf{98.60} & \textbf{96.32} &  -              &  -  \\
\bottomrule
\end{tabularx}
\end{table}

\begin{table}[t] 
\caption{Performance of VLM pre-training methods on the Visual Question Answering task; bold indicates the best performance. $\ast$: few-shot prediction.\label{VQA}}
\begin{tabularx}{\textwidth}{
>{\centering\arraybackslash}X
>{\centering\arraybackslash}X
>{\centering\arraybackslash}X
>{\centering\arraybackslash}X
}
\toprule
\textbf{Model}	& \textbf{RSVQA-LR}~\cite{RSVQA}	& \textbf{RSVQA-HR}~\cite{RSVQA}  & \textbf{RSIVQA}~\cite{RSIVQA} \\
\midrule
SkySenseGPT~\cite{skysensegpt} & \textbf{92.69}                   & 76.64 $^\ast$                             & - \\
RSGPT~\cite{rsgpt}             & 92.29                            & 92.00                               & - \\
GeoChat~\cite{geochat}         & 90.70                             & 72.30 $^\ast$                           & - \\
LHRS-Bot~\cite{Lhrs-bot}       & 89.19                            & \textbf{92.55}                   & - \\
H2RSVLM~\cite{h2rsvlm}         & 89.12                            & 74.35 $^\ast$                             & - \\
RS-LLaVA~\cite{RS-llava}       & 88.56                            &                     -             & - \\
SkyEyeGPT~\cite{skyeyegpt}     & 88.23                            & 86.87                            & - \\
EarthGPT~\cite{earthgpt}       &    -                              & 72.06 $^\ast$                        & - \\
%\midrule
SHRNet~\cite{A_Spatial_H} \textsuperscript{1}    & 85.85        & 85.39         & \textbf{84.46} \\
MQVQA~\cite{Multistep-VQA} \textsuperscript{1}     &  -  &   - & 82.18  \\
\bottomrule
\end{tabularx}
\noindent{\footnotesize{\textsuperscript{1} Only these are from the other part; the rest are conversational.}} 
%\noindent{\footnotesize{\textsuperscript{1} Tables may have a footer.}}
\end{table}

In Tables~\ref{VQA} and~\ref{IC}, we delve deeper into two frequently encountered visual--language tasks: Visual Question Answering (VQA) and Image Classification (IC). The results clearly illustrate that the conversational model, SkySenseGPT~\cite{skysensegpt}, maintains  exceptional performance across both tasks. Notably, none of the contrastive VLMs proved effective for the VQA task, further underscoring the advantages of conversational methods. Additionally, in the IC task, the prevalence of conversational VLMs exceeds that of their contrastive counterparts.

In conclusion, from Tables~\ref{tab:contrastive} and \ref{tab:conversational}, it is evident that a key architectural difference between conversational VLMs and contrastive VLMs lies in the use of pre-trained LLMs (e.g., Vicuna~\cite{Vicuna}, LLaMA~\cite{LLaMA}) in conversational VLMs. These LLMs process visual elements extracted by visual encoders and transformed by middleware, along with user queries or textual instructions. This enables conversational VLMs to handle more diverse and complex visual--language tasks, such as VQA and IC. As shown in Tables~\ref{VQA} and  \ref{IC}, conversational VLMs consistently outperform other types, further unlocking the potential of language to enhance reasoning in complex scenarios. Consequently, we advocate for adopting conversational model architectures as the preferred choice for future advancements in visual--language modeling.

\begin{table}[t] 
\caption{Performance %MDPI: Table should appear after first citation, please revise
 of VLM pre-training methods on image caption task; bold indicates the best performance. %EE: check meaning retained
 \label{IC}}
\begin{tabularx}{\textwidth}{
    >{\centering\arraybackslash}p{1.8cm}
    >{\raggedright\arraybackslash}X
    >{\centering\arraybackslash}p{1.1cm}
    >{\centering\arraybackslash}p{1.1cm}
    >{\centering\arraybackslash}p{1.1cm}
    >{\centering\arraybackslash}p{1.1cm}
    >{\centering\arraybackslash}p{1.3cm}
    >{\centering\arraybackslash}p{1.3cm}
    >{\centering\arraybackslash}p{1.3cm}
}
\toprule
\textbf{Dataset}	             & \textbf{Model}	         & \textbf{BLEU1}& \textbf{BLEU2}&\textbf{BLEU3} &\textbf{BLEU4} & \textbf{METEOR}& \textbf{ROUGEH}& \textbf{CIDEr}\\
\midrule
\multirow{5}{*}{\shortstack{UCM- \\ Captions~\cite{UCM}}}    & CRSR \textsuperscript{1}~\cite{cross-modal}    & 90.60           & 85.61          & 81.22          & 76.81          & \textbf{49.56} & \textbf{85.86} & \textbf{380.69} \\
                                 & SkyEyeGPT~\cite{skyeyegpt} & \textbf{90.71} & \textbf{85.69} & \textbf{81.56} & \textbf{78.41} & 46.24          & 79.49          & 236.75          \\
                                 & RS-LLaVA~\cite{RS-llava}   & 90.00             & 84.88          & 80.30           & 76.03          & 49.21          & 85.78          & 355.61          \\
                                  & RSGPT~\cite{rsgpt}         & 86.12          & 79.14          & 72.31          & 65.74          & 42.21          & 78.34          & 333.23          \\
                                 & RS-CapRet~\cite{RS-CapRet} & 84.30           & 77.90           & 72.20           & 67.00             & 47.20           & 81.70           & 354.80           \\    
                                 \midrule
\multirow{4}{*}{\shortstack{Syndney- \\ Captions~\cite{captiondatasets}} } & CRSR \textsuperscript{1}~\cite{cross-modal}    & 79.94          & 74.4           & 69.87          & 66.02          & 41.50           & 74.88          & \textbf{289.00}    \\
                                 & SkyEyeGPT~\cite{skyeyegpt} & \textbf{91.85} & \textbf{85.64} & \textbf{80.88} & \textbf{77.40}  & \textbf{46.62} & \textbf{77.74} & 181.06          \\
                                 & RSGPT~\cite{rsgpt}         & 82.26          & 75.28          & 68.57          & 62.23          & 41.37          & 74.77          & 273.08          \\         
                                 & RS-CapRet~\cite{RS-CapRet} & 78.70           & 70.00             & 62.80           & 56.40           & 38.80           & 70.70           & 239.20           \\ 
                                 \midrule
\multirow{4}{*}{RSICD~\cite{RSICD}}           & CRSR \textsuperscript{1}~\cite{cross-modal}    & 81.92          & 71.71          & 63.07          & 55.74          & \textbf{40.15} & \textbf{71.34} & \textbf{306.87} \\
                                 & SkyEyeGPT~\cite{skyeyegpt} & \textbf{87.33} & \textbf{77.70}  & \textbf{68.90}  & \textbf{61.99} & 36.23          & 63.54          & 89.37           \\
                                 & RSGPT~\cite{rsgpt}         & 70.32          & 54.23          & 44.02          & 36.83          & 30.1           & 53.34          & 102.94          \\
                                 & RS-CapRet~\cite{RS-CapRet} & 72.00             & 59.90           & 50.60           & 43.30           & 37.00             & 63.30           & 250.20           \\ 
                                 \midrule
\multirow{2}{*}{\shortstack{NWPU- \\ Captions~\cite{NWPUCaptions}}}   & RS-CapRet~\cite{RS-CapRet} & \textbf{87.10}  & \textbf{78.70}  & \textbf{71.70}  & \textbf{65.60}  & 43.60           & 77.60           & \textbf{192.90}  \\
                                 & EarthGPT~\cite{earthgpt}   & \textbf{87.10}  & \textbf{78.70}  & 71.60           & 65.50           & \textbf{44.50}  & \textbf{78.20}  & 192.60           \\
\bottomrule
\end{tabularx}
\noindent{\footnotesize{\textsuperscript{1} Only CRSR is contrastive; the rest are conversational.}} 
\end{table}

\section{Conclusions and Future Work }
\label{sec:con}
This review provides a systematic overview of Vision--Language Models (VLM) in the field of remote sensing. It introduces milestone works in the development of VLM itself, discusses the current research status of VLM in remote sensing, outlines the tasks involved, and offers a brief comparison of different VLM methods in various remote sensing applications. Overall, VLM has once again facilitated advancements in data processing technologies within the field of remote sensing. Current work has clearly demonstrated the potential of VLM applications in this area. However, there are still some shortcomings in current VLM technologies and their applications in remote sensing. Future work could consider the following directions:

\begin{itemize}
    %------David-----%
    \item \textbf{Addressing regression problems.}
    Currently, the mainstream approach in VLM typically involves using a tokenizer to vectorize text, aligning the visual information with word vectors, and then feeding these into LLMs. However, this method fails to capture numerical relationships effectively, which leads to problems in regression tasks. For instance, a number like `100' may be tokenized as `1', `00', or `100', causing loss of precision and affecting regression accuracy. Future work should focus on enhancing VLMs for regression tasks by addressing this limitation. 
    One promising direction is to develop specialized tokenizers for numerical values, which would ensure more accurate representation of numbers in text. Additionally, exploring the integration of task-specific regression heads into the existing VLM framework can help the model achieve a more efficient inference process. 
    For example, 
    %REO-VLM~\cite{REO-VLM} designs a regression head with a four-layer MLP-mixer like structure, which integrates knowledge embeddings generated by the LLM hidden tokens with complex visual tokens. 
    REO-VLM~\cite{REO-VLM} designs a regression head with a four-layer MLP-mixer-like structure that integrates knowledge embeddings derived from LLM hidden tokens with complex visual tokens.
    This innovative approach enables VLMs to effectively handle regression tasks such as Above-Ground Biomass (AGB)~\cite{agb1, agb2} estimation in Earth observation (EO) applications.
    Task-specific regression heads enable the model to directly output data in formats that meet the specific requirements of the task, rather than being limited to regression on text tokens. Moreover, these heads can be dynamically adjusted based on the needs of different tasks, making the regression output more accurate and efficient. Introducing the Mixture of Experts (MOE) architecture to improve the adaptability and accuracy of regression tasks is also an excellent choice. When deployed in VLMs, it can use a gating mechanism to dynamically match the appropriate expert modules for different tasks, enabling the model to effectively handle multi-task learning with sufficient capability.

    \item \textbf{Aligning with the structural characteristics of remote sensing images.} 
    % 高光谱、SAR (已与大伟写的整合)(审稿人4 comment 6) gtwei
    Current VLMs in remote sensing largely adopt frameworks developed for conventional computer vision tasks and rely on models pre-trained exclusively on RGB imagery, thereby failing to fully account for the distinctive structural and spectral characteristics of remote sensing data. This limitation is particularly evident for modalities such as SAR and HSI, where crucial features remain poorly captured by general-purpose extractors. As a result, while existing VLMs may perform reasonably well on RGB data, they struggle with SAR and HSI inputs and often depend on superficial fine-tuning with limited datasets. To address this gap, future research should emphasize both the design of specialized feature extractors tailored to the complexities of multispectral and SAR data, and the development of algorithms capable of handling multidimensional inputs. REO-VLM~\cite{REO-VLM} makes effective attempts by introducing spectral recombination and pseudo-RGB strategies. It splits the high-dimensional MS image into multiple three-channel images and generates a third channel for the SAR images, which typically contain two polarization channels, thus creating a pseudo-RGB representation compatible with the pre-trained visual encoder.
    Expanding and diversifying remote sensing datasets to include large-scale image--text pairs from multiple modalities (e.g., RGB, SAR, HSI, LiDAR) will be essential for training VLMs that can achieve robust visual--text alignment across all data types. By integrating richer and more comprehensive feature representations, as well as exploring data fusion strategies, it will be possible to enhance VLM performance in a wide range of complex remote sensing applications, including land-use classification, urban monitoring, and disaster response.
    % 高光谱、SAR (已与大伟写的整合) gtwei
    % Current VLM efforts in the remote sensing field largely follow the overarching framework established in the computer vision domain, without fully leveraging the unique characteristics of remote sensing images. For instance, remote sensing images come in various types, such as SAR images, HSI, and radar images, each with distinct features. Designing feature extractors that cater to these data characteristics is essential to maximize the advantages of images. However, in current remote sensing VLM work, common methods include manual annotation to construct datasets or using low-cost RGB images as bridges to align different RSI images, which do not fully align with remote sensing images. Additionally, there is a lack of research considering the fusion of different modalities of remote sensing images to enhance the performance of VLM.
    \item \textbf{Multimodal output.} 
    % 多任务输出(已与大伟写的整合)(审稿人4 comment 7) gtwei
    Early methods, such as classification models, were constrained by their output format and thus limited to a single task. Although most current VLMs, which produce text outputs, have broadened their scope and can effectively handle numerous tasks, they still fall short for dense prediction tasks like segmentation and Change Detection. To overcome this limitation, future research should focus on enabling VLMs to produce truly multimodal outputs—such as images, videos, and even 3D data. 
    Recent studies have begun exploring the enhancement of traditional proprietary models by adding a text branch as an auxiliary component, thereby fostering richer multimodal feature interaction and improving image feature extraction and interpretation while preserving existing output heads. Building on this idea, another promising direction is to use a VLM as an agent that interfaces with multiple specialized model output heads and activates the most suitable one based on the task identified, thus effectively combining the reasoning power of LLMs with the expertise of dedicated models. In addition, a recent insightful approach is to use next-token prediction to achieve multimodal output. For example, Emu3~\cite{emu3} adopts a completely next-token prediction-based architecture, avoiding the complex diffusion models or combinatorial methods used in the past. It innovatively uses visual tokens as part of the output to generate images and videos, and outperforms well-known open-source models.%For example, to handle regression tasks, VLMs can be equipped with a dedicated regression head that, when engaged for such tasks, generates numerical results to fill predefined placeholders within the VLM’s output, thereby seamlessly integrating regression-based predictions into the multimodal framework.
    % 多任务输出(已与大伟写的整合) gtwei
    %Existing VLM models in the remote sensing field typically output only text. While they can handle traditional tasks like image classification, they struggle with dense prediction tasks such as segmentation, where text output is insufficient. Therefore, researching multimodal outputs, allowing VLM models to also output images or even videos, is a highly valuable direction. This could facilitate the integration of various application needs within the remote sensing domain into a unified framework.
    \item \textbf{Multitemporal Regression and Trend Inference.}
    Current VLM research in the remote sensing field often focuses solely on static images, overlooking the potential of multitemporal data. Since remote sensing data can capture information that changes over time, future research should explore how to extend VLMs to handle multitemporal regression tasks. This involves not only spatial feature recognition but also the ability to infer temporal trends and dynamics. A key research direction is to develop methods that utilize historical remote sensing data to analyze and predict long-term environmental or ecological changes. This is particularly important for applications such as climate change monitoring, land-use change prediction, and ecosystem management, where temporal patterns play a crucial role. Remote sensing images with temporal continuity can be encoded as sequences, providing multitemporal data as input for training VLMs. By adding multitemporal regression capabilities to VLMs, the model could provide more context-aware predictions, improving the accuracy of long-term environmental assessments.
    
\end{itemize}

%Bibliography
\bibliographystyle{unsrt}  
\bibliography{reference}

\end{document}